\documentclass{article} % For LaTeX2e
\usepackage[preprint]{colm2025_conference}

\usepackage{microtype}
\usepackage{hyperref}
\usepackage{url}
\usepackage{booktabs}
\usepackage{latexsym}
\usepackage{amsmath}
\usepackage{amssymb}
\usepackage{xcolor} 
\usepackage{mathtools}
\usepackage{amsthm}
\usepackage{algorithm}
\usepackage{algpseudocode}
\usepackage{multirow}
\usepackage{booktabs}
\usepackage{lineno}

\definecolor{darkblue}{rgb}{0, 0, 0.5}
\hypersetup{colorlinks=true, citecolor=darkblue, linkcolor=darkblue, urlcolor=darkblue}
\newcommand{\method}{RaM }

\title{Rational Metareasoning for Large Language Models}

\author{
    \textbf{Nicolò De Sabbata}{\hspace{.1em}}
    \quad
    \textbf{Ted Sumers}{\hspace{.1em}}
    \quad
    \textbf{Badr AlKhamissi}{\hspace{.1em}}
    \quad
    \vspace{.2em}\\
    \textbf{Antoine Bosselut}{\hspace{.1em}}
    \quad
    \textbf{Thomas L. Griffiths}{\hspace{.1em}}
    \quad
    \vspace{.5em}\\
    EPFL
    \vspace{.5em}\\
    \texttt{jcheng71@jhu.edu}
}
\author{
    Nicolò De Sabbata$^{1, 2}$\thanks{Now at Apple},\quad Theodore R. Sumers$^{3}$,\quad Badr AlKhamissi$^{2}$ \\
    \textbf{Antoine Bosselut$^{2}$,\quad Thomas L. Griffiths$^{1}$}\\
    $^1$Princeton University,\quad $^2$EPFL,\quad $^3$Anthropic
}

\begin{document}

\ifcolmsubmission
\linenumbers
\fi

\maketitle

\begin{abstract}

% Many recent LLM-based reasoning methods deploy additional inference-time computation to boost performance on complex tasks.

Recent approaches for leveraging large language models (LLMs) for reasoning often rely on additional inference-time computation to tackle complex tasks. However, as LLMs grow in size, the associated inference costs are becoming increasingly prohibitive. To address this, we introduce \textbf{\method} (\textbf{Ra}tional \textbf{M}etareasoning): a cost-aware reasoning approach inspired by computational models of metareasoning in cognitive science. Our method trains LLMs to use intermediate reasoning steps selectively when they are likely to be beneficial. We first design a reward function that incorporates the Value of Computation by penalizing unnecessary reasoning, and then use it within an Expert Iteration framework to optimize the trade-off between performance and efficiency. Compared to few-shot chain-of-thought prompting and STaR, our method significantly reduces inference costs (23-45\% fewer tokens generated across three models) while maintaining or increasing task performance across diverse datasets.

\end{abstract}

%%%%%%%%%%%%%%%%%%%%%%%%%%%%%%%%%%%%%%%%%%%%%%%%%%%%%%%%%%%%
\section{Introduction}
%%%%%%%%%%%%%%%%%%%%%%%%%%%%%%%%%%%%%%%%%%%%%%%%%%%%%%%%%%%%

Large language models (LLMs) rely on substantial computational power to handle complex problems \citep{openai2024gpt4technicalreport, chowdhery2022palmscalinglanguagemodeling, DEVRIES20232191}. While initial studies mostly focused on the cost of training \citep{Verdecchia2023}, widespread use of LLMs  has made inference-time costs an increasingly important factor. Moreover, there is a fundamental tension between inference cost and task performance: although chain-of-thought prompting \citep[CoT;][]{wei2023chainofthoughtpromptingelicitsreasoning, kojima2023largelanguagemodelszeroshot} and similar approaches improve task performance, they also substantially increase inference costs \citep{snell2024scalingllmtesttimecompute}.

This trend has recently spiked with the development of Large Reasoning Models \citep{o1systemcard, deepseekai2025deepseekr1incentivizingreasoningcapability}, instruction-tuned LLMs which are typically post-trained using reinforcement learning (RL) with outcome-based rewards to produce long chains of thought before each response. It is also worth noting that these approaches are not inherently \emph{adaptive}: inference optimization \citep{wan2024efficientlargelanguagemodels} and existing CoT training methods often tend to raise or lower the inference cost on {\em all} queries, regardless of task complexity, or require the user to specify a budget up front \citep{claude37sonnet}.

In stark contrast to this static tradeoff, humans are able to adaptively allocate computational resources based on task difficulty \citep{Kahneman, RUSSELL1997,Lieder2017}. In this work, we draw inspiration from \emph{rational metareasoning} -- literally, reasoning about reasoning -- a concept originally from the artificial intelligence literature \citep{Russell1991} that has been used to explain how humans adaptively manage computational resources \citep{Lieder2017, Lieder2018, Griffiths2019}.

Building on this, we develop a novel reward function based on the Value of Computation \citep[VOC;][]{Russell1991}, which formalizes the trade-off between inference cost and task performance. We adopt an iterative reinforcement learning process inspired by the Expert Iteration algorithm \citep{anthony2017thinkingfastslowdeep}. In each iteration, we generate multiple reasoning chains for each question. These reasoning chains are ranked using the reward function, and the dataset is filtered to retain only the best reasoning chain for each question. The model is then fine-tuned using this filtered dataset. However, unlike previous applications of Expert Iteration to LLMs \citep{zelikman2022starbootstrappingreasoningreasoning, havrilla2024teachinglargelanguagemodels}, which filter generated examples solely based on the correctness of the final answer, our method optimizes for both correctness \emph{and the cost} of the reasoning process.

We evaluated the effectiveness of our solution across a diverse set of tasks, from scientific knowledge \citep[ARC;][]{clark2018thinksolvedquestionanswering} to commonsense reasoning \citep[CommonsenseQA;][]{talmor2019commonsenseqaquestionansweringchallenge}, mathematical problem solving \citep[GSM8K;][]{cobbe2021trainingverifierssolvemath}, and logical deductive reasoning \citep[ProofWriter;][]{tafjord2021proofwritergeneratingimplicationsproofs}. Additionally, we assess the out-of-domain generalization on MMLU \citep{hendrycks2021measuringmassivemultitasklanguage}, a multitask benchmark. Our approach achieves a substantial reduction in generated tokens (35-42\% compared to few-shot prompting, and 23-32\% compared to STaR) while matching or increasing performance. Thus, we make the following contributions:
\begin{enumerate}
\item{We employ rational metareasoning to optimize the tradeoff between inference cost and performance of LLMs on reasoning tasks.}
\item{We formalize a novel reward function inspired by the Value of Computation (VOC) and integrate it into LLM training.}
\item{We empirically demonstrate that rational metareasoning improves task performance at lower inference costs (23-42\% fewer tokens on average) across various datasets and reasoning tasks.}
\end{enumerate}

%%%%%%%%%%%%%%%%%%%%%%%%%%%%%%%%%%%%%%%%%%%%%%%%%%%%%%%%%%%%
\section{Rational Metareasoning}
%%%%%%%%%%%%%%%%%%%%%%%%%%%%%%%%%%%%%%%%%%%%%%%%%%%%%%%%%%%%

Humans have limited time and cognitive resources \citep{Griffiths2019, Griffiths2020}. We face diverse challenges requiring different approaches: avoiding a sudden obstacle when driving needs quick, intuitive thinking, whereas selecting a retirement investment strategy requires slow, deliberate reasoning \citep{Kahneman}. Rational metareasoning \citep{Russell1991} captures this by suggesting agents should adapt their reasoning based on the problem at hand.

Intuitively, while reasoning solves a problem, metareasoning solves the problem of \emph{how} to solve a problem: deciding which computations to perform while problem-solving. The essence of rational metareasoning is calculating the value of computation \citep[VOC;][]{Russell1991} for each potential computation. The VOC balances the benefit of computation $c$ (the expected increase in the agent's  utility) against its cost (usually time or energy). 

To formalize this, agents are assumed to have some internal belief state $b \in \mathcal{B}$, which determines their expectation about the value of each action $a \in \mathcal{A}$: $\mathbb{E}[U(a)|b]$. A rational agent would simply choose the highest-value action: $a^* = \text{argmax}_{a\in\mathcal{A}}{[U(a)|b]}$. However, this picture becomes more complex if an agent can perform computation to change their belief state before choosing an action. Each computation $c\in\mathcal{C}$ updates the agent's belief to $b'$ with probability $P(b'| c)$, which in turn affects their beliefs about the value of actions, and incurs an associated cost ($\text{cost}(c)$). The VOC quantifies the value of performing computation $c$ given a starting belief state $b$,

\begin{equation} \label{eq:1}
VOC(c, b) = \mathbb{E}_{P(b'| c)}[\max_{a'}\mathbb{E}[U(a')|b']-\max_a\mathbb{E}[U(a)|b]]-\text{cost}(c).
\end{equation}

Thus, a meta-rational agent should pursue the computation $c^*$ with the highest VOC: $c^* = \text{argmax}_{c\in\mathcal{C}}VOC(c, b)$. If no computation has positive VOC, the agent should stop thinking and act in the world. Rational metareasoning can explain how humans allocate cognitive resources in various tasks \citep{Lieder2017, Lieder2018, callaway2018learning, callaway2021fixation, callaway2022rational, russek2022time}.

%%%%%%%%%%%%%%%%%%%%%%%%%%%%%%%%%%%%%%%%%%%%%%%%%%%%%%%%%%%%
\section{Rational Metareasoning with Large Language Models}
%%%%%%%%%%%%%%%%%%%%%%%%%%%%%%%%%%%%%%%%%%%%%%%%%%%%%%%%%%%%

To achieve an optimal balance between performance and efficiency, our approach introduces a new VOC-inspired reward function (Eq. \ref{eq:2}) into Expert Iteration \citep{anthony2017thinkingfastslowdeep, zelikman2022starbootstrappingreasoningreasoning}, fine-tuning a LLM to produce reasoning chains adaptively. % depending on task difficulty.

\begin{figure}
\centering
\includegraphics[width=0.85\textwidth]{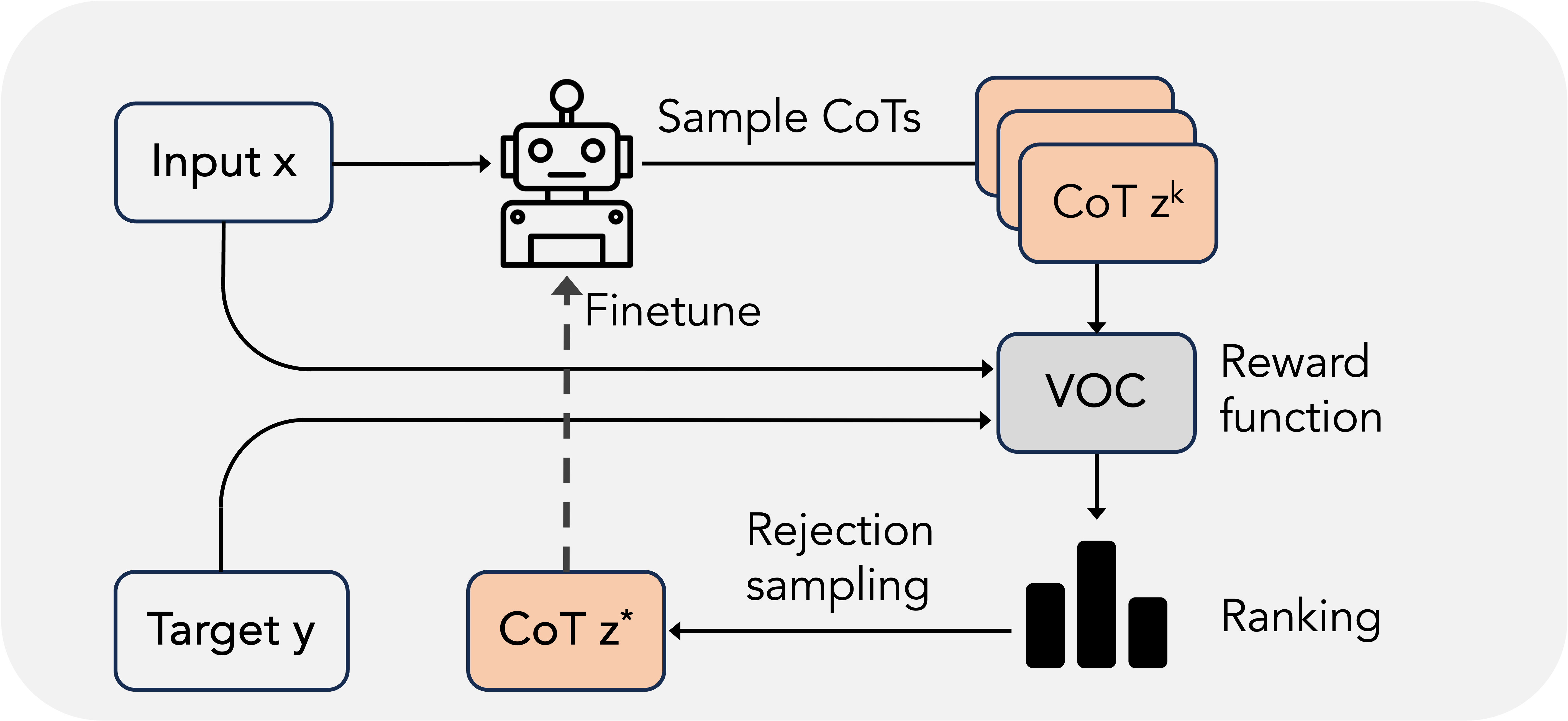}
\begin{small}
\caption{\textbf{RaM training}. We iteratively generate reasoning chains using the current policy, score and filter them to approximate the optimal policy, and then finetune the base policy.}
\end{small}
\label{fig:mr_algo}
\end{figure}

\subsection{Reward modeling} \label{reward_modeling}

Chain-of-thought prompting encourages LLMs to generate an intermediate output -- a ``chain of thought'' -- prior to producing the answer to a question \citep{wei2023chainofthoughtpromptingelicitsreasoning, kojima2023largelanguagemodelszeroshot}.
We define the reward of a chain of thought as the difference between its utility and its cost, 
\begin{equation} \label{eq:2}
\mathcal{R}_{\pi}(x, y, z) = \mathcal{U}_{\pi}(z|x, y) - \mathcal{C}(z) 
\end{equation}
where $x$ denotes the input for the task, $z$ represents the chain of thought, and $y$ is the target solution. The utility of the chain of thought is represented by $\mathcal{U}_{\pi}(z|x, y)$, and the cost of the intermediate computations is denoted by $\mathcal{C}(z)$. Equation \ref{eq:2} mirrors the VOC Equation \ref{eq:1}: here, individual reasoning tokens correspond to intermediate computations $ c $, making the reasoning chain $ z $ a sequence of computations $ c $, while the actions $ a \in \mathcal{A} $ map to the potential outputs $y \in \mathcal{Y}$ of the language model. In the context of LLMs, utility quantifies the increase in the likelihood of generating the target sequence $y$ when the chain of thought $z$ is added to the input $x$, under the policy $\pi$:
\begin{equation} \label{eq:3}
    \mathcal{U}_{\pi}(z|x, y) = \log{\pi_\theta(y|z, x)} - \log{\pi_\theta(y|x)}.
\end{equation}
Specifically, $\pi_{\theta}(y|z, x)$ indicates the probability of generating the target sequence $y$ given both the chain of thought $z$ and the input $x$, while $\pi_{\theta}(y|x)$ denotes the probability of generating $y$ with only the input $x$. With respect to Equation \ref{eq:1}, the language model’s initial belief about the value of actions or outputs is described by $\pi_\theta(y \mid x)$ (from Equation \ref{eq:3}), whereas its final belief after generating a sequence of computations or tokens (a chain of thought $z$) is described by $\pi_\theta(y \mid z, x)$. The cost is directly proportional to the number of tokens in the chain of thought $l(z)$:
\begin{equation}
\mathcal{C}(z) = \gamma\cdot l(z).
\end{equation}
The hyperparameter $\gamma$ scales the cost and utility to the same magnitude. A key benefit of this reward function is that it is parameterized by the same weights $\theta$ as the generative policy $\pi_\theta$, eliminating the need for an external reward model. This allows for direct estimation of the utility of a reasoning chain using the policy itself. 

Finally, it’s worth noting that while standard reward models for RLHF bias the policy to produce longer reasoning chains \citep{singhal2024longwaygoinvestigating}, thus necessitating appropriate corrections \citep{chen2024odindisentangledrewardmitigates, park2024disentanglinglengthqualitydirect} to correct for reward hacking, our reward model aims for and achieves the opposite effect.

\subsection{Metareasoning Training}

We demonstrate the effectiveness of our reward model using a variation of the Expert Iteration algorithm \citep[EI, ][]{anthony2017thinkingfastslowdeep}. EI is known for its sample efficiency and strong performance on reasoning tasks \citep{havrilla2024teachinglargelanguagemodels, zelikman2022starbootstrappingreasoningreasoning}. As an example of an online reinforcement learning algorithm, EI involves both exploration and policy improvement phases, with the policy $\pi_\theta$ being updated using data from the exploration phase (as can be seen in Algorithm \ref{alg:rm} and Figure \ref{fig:mr_algo}).

Initially, in the exploration phase, we approximate the optimal policy $\hat{\pi}^*$ (whose computations maximize the VOC reward) by using rejection sampling on our student policy $\pi_\theta$. In particular, we begin with a pretrained language model $\pi_{\theta}$ and an initial dataset of problems $x$ along with their corresponding correct final answers $y$: $\mathcal{D} = \{(x_i, y_i)\}_{i=1}^D$. Following prior work in online RL \citep{tang2024understandingperformancegaponline}, we utilize the model itself to generate the reasoning chains. Since we start with a pretrained model, we guide the model generation using few-shot prompting in the first iterations. Specifically, we concatenate a small set of examples, denoted as $\mathcal{P}$, each containing intermediate reasoning chains $z$, to each example in $\mathcal{D}$. During training and evaluation, we remove the few shot examples. 

For each task $\tau_i=(x_i, y_i)$ in the original dataset $\mathcal{D}$, we generate $K$ reasoning chains: $\hat \tau_{i} = \{(x_i, z_{k,i}, y_i)\}_{k=1}^K$. Then, we evaluate them using our reward function \(\mathcal{R}_\pi\) (Section ~\ref{reward_modeling}), and compute the advantage of each chain by subtracting the average reward for that question: $a_{i,k} = r_{i,k} - \frac{1}{K} \sum_{k'=1}^K r_{i,k'}$. 

Using these advantage scores, we perform rejection sampling (by discarding reasoning chains with negative advantage) to construct the dataset of optimal trajectories $(\mathcal{D}_1^*)$. Finally, we distill the selected rollouts into a policy \(\pi_1\) via standard cross-entropy loss. This process can be iteratively repeated to refine the policy $\pi_n$ on the dataset $\mathcal{D}_n^*$. In closed-answer settings with verifiable answers, the reasoning chains that lead to incorrect answers can also be directly discarded, to produce a higher quality dataset.

\vspace{-2mm}

\begin{algorithm}
\caption{Rational Metareasoning Training}\label{alg:rm}
	    \hspace*{\algorithmicindent} \textbf{Input} $\pi$: a pretrained LLM; dataset $\mathcal{D} = \{(x_i, y_i)\}_{i = 1}^D$
	\begin{algorithmic}[1]
        \State $\pi_0 \leftarrow \pi$ \Comment{\textcolor{gray}{Copy the original model}} 
		\For {$n$ \textbf{in} $1...N$} 
        %\State ${\mathcal{D}_n} \leftarrow \{\mathcal{T}\subseteq \mathcal{D} \mid |\mathcal{T}| = T\} \cup \mathcal{D}_{n-1} $ \Comment{\textcolor{gray}{Sample batch from dataset}}
            \For {$k$ \textbf{in} $1...K$}
		    \State $({z}_{i, k}, {y}_{i, k}) \leftarrow \pi_{n - 1}(x_i)\quad \forall i \in [1, D]$
            \Comment{\textcolor{gray}{Sample reasoning chains}}
            \EndFor
            \State $r_{i,k} \leftarrow \mathcal{R}_{\pi_{n-1}}(x_i, y_i, z_{i, k}) \quad \forall i, k ( i \in [1, D], k \in [1, K])$
            \Comment{\textcolor{gray} {Compute rewards}}
            \State $a_{i,k} \leftarrow  r_{i,k} - \frac{1}{K} \sum_{k'=1}^K r_{i,k'} \quad \forall i, k ( i \in [1, D], k \in [1, K])$
            \Comment{\textcolor{gray} {Compute advantages}}
            \State $\hat{Z}_i \leftarrow \{\, z_{i, k} \mid a_{i,k} > 0 \,\}$
            \Comment{\textcolor{gray} {Select best reasoning chains}}
            \State $\mathcal{D}_n^* \leftarrow \{\, (x_i, \hat{z}_{i,k} , y_i) \mid \hat{z}_{i,k} \in \hat{Z}_i, \, i \in [1, D] \,\}$ 
            \Comment{\textcolor{gray} {Create the optimal dataset}}
            \State $\pi_n \leftarrow \text{train}(\pi, \mathcal{D}_n^* )$
            \Comment{\textcolor{gray} {Finetune the original model on the optimal solutions}}
		\EndFor
	\end{algorithmic}
\end{algorithm}
\vspace{-1em}

%%%%%%%%%%%%%%%%%%%%%%%%%%%%%%%%%%%%%%%%%%%%%%%%%%%%%%%%%%%%
\section{Experiments}
%%%%%%%%%%%%%%%%%%%%%%%%%%%%%%%%%%%%%%%%%%%%%%%%%%%%%%%%%%%%
%We now detail the datasets (Sec.~\ref{datasets}), baselines (Sec.~\ref{baselines}), and training (Sec.~\ref{training}) used to evaluate our method. 

\subsection{Datasets} \label{datasets}

To characterize the generality of our method, we applied it to a diverse range of datasets and reasoning tasks. We constructed our training set by combining the training sets from the following datasets into one dataset $\mathcal{D}$ and then evaluated the model on all corresponding public test sets $\mathcal{T}$.  To facilitate comparison with baselines that use the correctness of the answer as a reward, we limited the training to datasets with verifiable answers, although this is not strictly required by our method. We used four datasets:

%\begin{itemize}
    \textbf{ARC} \citep{clark2018thinksolvedquestionanswering}. The AI2 Reasoning Challenge (ARC) dataset comprises grade-school science questions, designed to evaluate the  capability to apply scientific knowledge.
    
    \textbf{CommonsenseQA} \citep{talmor2019commonsenseqaquestionansweringchallenge}. This dataset is centered on commonsense question answering. It leverages implicit human knowledge and testing everyday reasoning.
    
    \textbf{GSM8K} \citep{cobbe2021trainingverifierssolvemath}. This dataset includes a variety of linguistically diverse grade-school math word problems. It assesses proficiency in solving mathematical problems that require comprehension and application of arithmetic reasoning.
    
    \textbf{ProofWriter} \citep{tafjord2021proofwritergeneratingimplicationsproofs}. This dataset assesses logical deductive reasoning by asking the model to determine if a conclusion follows from premises presented in natural language.
%\end{itemize}

These datasets have very different train split sizes. To ensure fairness and balance between the datasets, and to manage computational costs, we composed our training mixture by sampling 1024 random samples from each of the training sets (for a total of 4096 samples). We then evaluated the model on the public test set of each dataset. To further assess the generalization of our approach, we conducted out-of-distribution testing on \textbf{MMLU-CF} \citep{zhao2024mmlucfcontaminationfreemultitasklanguage}, a contamination free and more challenging variant of the original MMLU dataset \citep{hendrycks2021measuringmassivemultitasklanguage}. This dataset consists of 10,000 multiple-choice questions from various branches of knowledge.

\subsection{Baselines} \label{baselines}

We illustrate the advantages of our model by comparing its performance to two types of prompting strategies: \textbf{Direct prompting}, where the model is required to provide an immediate answer, and \textbf{Chain of Thought prompting (CoT)}, where the model is encouraged to reason through the problem step-by-step before arriving at a solution. Since we are using pretrained models which are not specifically trained for instruction following, we provide five few-shot examples for each task from the unused portion of the training dataset. These are the same examples that are used to guide the model training during the first iteration. In addition to these prompting methods, we adopt as a finetuning baseline the most common reasoning bootstrapping method, \textbf{STaR} (Self-Taught Reasoner; \citealt{zelikman2022starbootstrappingreasoningreasoning}), which also uses a variation of Expert Iteration. In particular, this method was the first to adopt outcome-based rewards to enable models to self-improve: given a set of problems and solutions, it trains the model iteratively on reasoning chains that lead to the correct solution.

As an upper bound, we also tested the instruction-tuned version of the models, which have been fine-tuned on a massive amount of annotated chain-of-thought data, followed by RLHF \citep{dubey2024llama3herdmodels}. Finally, while CoT prompting may not yield optimal trajectories, more advanced methods \citep{yao2023treethoughtsdeliberateproblem, zheng2024stepbackevokingreasoning, madaan2023selfrefineiterativerefinementselffeedback} often increase sequence lengths substantially, and are therefore less efficient. We focus on CoT for its simplicity -- reducing reasoning tokens while gaining performance for CoT suggests similar gains for more complex approaches.

\subsection{Training Details}  \label{training}

For our experiments, we use Meta Llama-3.2-3B and Llama-3.1-8B \citep{dubey2024llama3herdmodels} as the pretrained base models. We have chosen $\gamma=0.1$ to align the distributions of costs and rewards more closely, although we have found our method to be robust to small variations in the choice of this hyperparameter. We sample $K=4$ reasoning chains for each question, using a temperature $t$ of 0.5 and a $top_p$ value of 0.9. These parameters are chosen to balance exploration and exploitation, allowing us to generate diverse yet relevant reasoning chains.

For greater efficiency during training, instead of using the entire dataset from the start, we begin by sampling a dataset \(\mathcal{D}_n\) of size \(T = 1024\) from the union of the four training datasets \(\mathcal{D}\) described in \ref{datasets}. We then progressively increase the dataset size by $T$ with each subsequent iteration (until it reaches the size of $\mathcal{D}$) allowing the model to encounter new examples gradually. In the self-supervised fine-tuning step, we use a batch size of 16 and a learning rate of 1e-5. We execute five iterations of the training algorithm. We believe that further improvements are possible through a more comprehensive hyperparameter search; however, due to computational constraints, we leave this for future work. Finally, we evaluate all models using greedy decoding to ensure consistent and deterministic output generation. We use pattern matching techniques to extract the answers; an exact match between the generated answer and the ground truth is considered correct.

\vspace{-0.5em}
\section{Results} \label{results}

\subsection{Performance vs Cost}\label{perf_vs_cost}
\begin{figure}
\vspace{-0.5em}
\includegraphics[width=\textwidth]{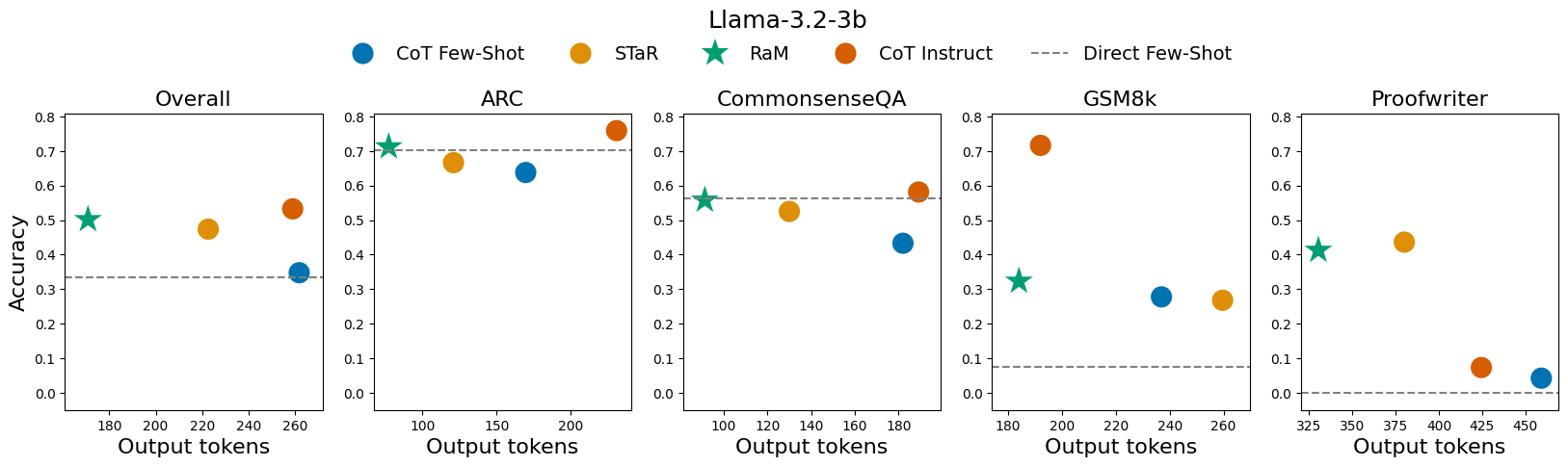}
\includegraphics[width=\textwidth]{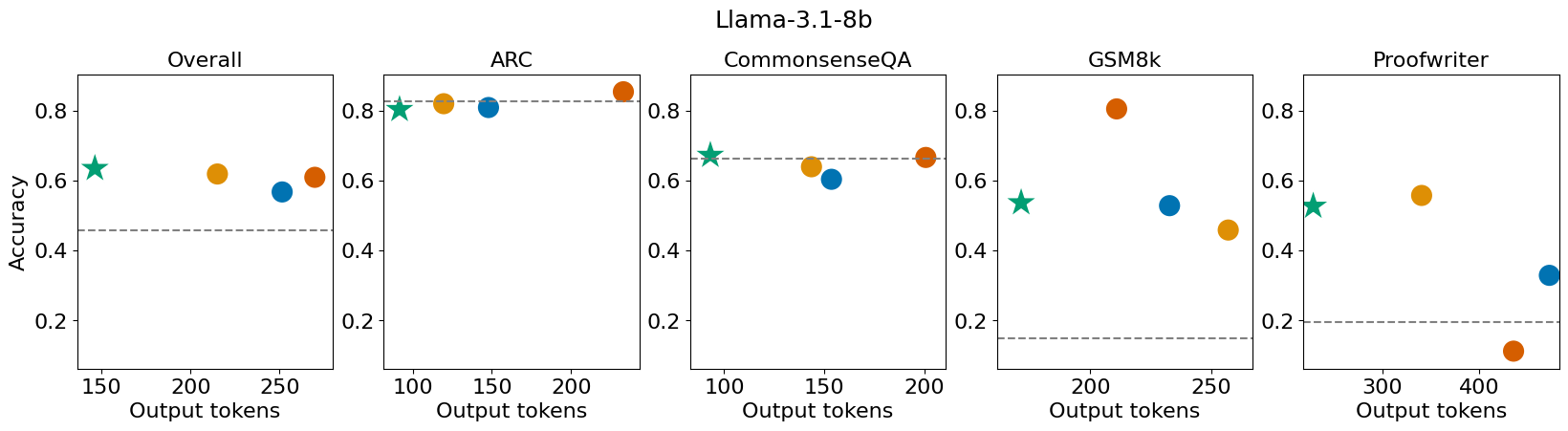}
\vspace{-2em}
\begin{small}
\caption{\textbf{Cost and performance}. Accuracy is plotted against the output tokens. Our method, RaM, eliminates the need for few-shot prompting (reducing input
tokens) and trains the model to use fewer reasoning tokens than STaR. }
\label{fig:rm_results_new}
\end{small}
\vspace{-1em}
\end{figure}

We first evaluate our approach against baselines (Sec.~\ref{baselines}) across several datasets (Sec.~\ref{datasets}). Our key criteria are \emph{performance} (measured by rescaled accuracy) and \emph{cost} (measured by the number of input and output tokens). Our experiments confirm that across all models and datasets, our training approach reduces cost while matching or improving performance (see Table~\ref{table:comparison} for results averaged across datasets; Tables~\ref{table:accuracies} and~\ref{table:output_lengths} for per-dataset results; and Appendix~\ref{qualitative_examples} for example reasoning chains). Fig.~\ref{fig:rm_results_new} shows these results for all models.

\begin{table}
\centering
\begin{small}
\begin{tabular}{@{}clccc@{}}
\toprule
\textbf{Model} & \textbf{Method} & \textbf{Accuracy (\%) $\uparrow$} & \textbf{\# Input Tokens $\downarrow$} & \textbf{\# Output Tokens $\downarrow$} \\ \midrule
\multirow{4}{*}{Llama-3.2-3B}
& Direct Few-Shot 
   & 33.5 ($\pm$ 1.4)
   & 291.7 ($\pm$ 0.9)
   & 0.0 ($\pm$ 0.0) 
   \\
   \cmidrule(lr){2-5}
 & CoT Few-Shot 
   & 34.8 ($\pm$ 1.8)
   & 1080.5 ($\pm$ 0.9)
   & 261.9 ($\pm$ 4.8)
   \\
 & STaR
   & 47.4 ($\pm$ 2.0)
   & \textbf{73.2} ($\pm$ 0.9)
   & 222.7 ($\pm$ 3.5)
   \\
 & \method
   & \textbf{50.1} ($\pm$ 1.8)
   & \textbf{73.2} ($\pm$ 0.8)
   & \textbf{170.8} ($\pm$ 2.9)
 \\ \midrule
 \multirow{1}{*}{Llama-3.2-3B-Instruct}
 & CoT Prompting & \textbf{53.2} ($\pm$ 1.8)
   & 157.2 ($\pm$ 0.9)
   & 259.1 ($\pm$ 3.4)
   \\ \midrule
\multirow{4}{*}{Llama-3.1-8B}
 & Direct Few-Shot 
   & 45.8 ($\pm$ 1.5)
   & 291.7 ($\pm$ 0.9)
   & 0.0 ($\pm$ 0.0)
   \\
   \cmidrule(lr){2-5}
 & CoT Few-Shot 
   & 56.4 ($\pm$ 1.9)
   & 1080.5 ($\pm$ 0.9)
   & 253.6 ($\pm$ 4.2)
   \\
 & STaR
   & \textbf{61.9} ($\pm$ 1.6)
   & 73.2 ($\pm$ 0.9)
   & 215.1 ($\pm$ 3.3)
   \\
 & \method
   & \textbf{63.5} ($\pm$ 1.7)
   & 73.2 ($\pm$ 0.9)
   & \textbf{146.2} ($\pm$ 3.3) 
 \\ \midrule
 \multirow{1}{*}{Llama-3.1-8B-Instruct}
 & CoT Prompting & \textbf{60.9} ($\pm$ 1.7)
   & 157.2 ($\pm$ 0.9)
   & 269.9 ($\pm$ 3.8)
   \\ \bottomrule
\end{tabular}
\end{small}
\vspace{-1em}
\begin{small}
\caption{Comparison of different methods based on rescaled accuracy and length metrics, averaged across datasets (means with 95\% confidence intervals; bold indicates best performing approaches with overlapping 95\% intervals). \method achieves better performance while using significantly fewer input and output tokens compared to STaR or CoT Few-Shot.}
\label{table:comparison}
\end{small}
\vspace{-1em}
\end{table}

We first consider the performance of our baselines: CoT Few-Shot prompting uses a large number of input and output tokens, but yields reasonable performance.  Direct Few-Shot prompting uses fewer input (and far fewer output) tokens, but yields poor performance on reasoning-intensive datasets (GSM8K and Proofwriter, see Table \ref{table:accuracies}). STaR improves on these approaches, using significantly fewer tokens while achieving comparable or superior performance.

\method further improves the cost-performance tradeoff by generating 23-32\% fewer tokens on average compared to STaR, generally with higher accuracy. Interestingly, our findings align with the conclusions of a meta-analysis of over 100 papers \citep{sprague2024cotcotchainofthoughthelps}, in finding that explicit chains help mainly on maths or logic tasks and bring marginal gains elsewhere.

Finally, it is worth noting that while the performance of the instruction-tuned model (CoT Instruct) is generally higher than that of the pretrained model, the length of the reasoning chains was generally much higher and not meaningfully adaptive to task complexity.

\subsection{Adaptive computation}
Section~\ref{perf_vs_cost} demonstrates that our method reduces computational costs \emph{on average}. But does it actually teach models to reason \emph{adaptively} (by adjusting reasoning to match task complexity), or just to reason \emph{less}? To address this question, we first divided our test set $\mathcal{T}$ based on whether or not reasoning was needed to solve the task. In particular, we split the data based on whether Direct Few-Shot obtained the correct answer (``easy split'') or not (``hard split''). Adaptive methods should use less computation to solve the easy problems. We can empirically compare the results across methods for these two data splits.

As shown in Table \ref{table:comparison_length}, all models and methods are able to differentiate between hard and easy problems, generating fewer tokens on easier problems. The instruction-tuned model (CoT Instruct) did not effectively adapt response lengths to align with task complexity in our experiments. While STaR seems to improve this ratio, \method increases the difference in reasoning between hard and easy problems, achieving a length reduction of up to 50.3\% on Llama-3.2-3B. This indicates that \method trains models to reason adaptively, helping them recognize when detailed reasoning is necessary and when a shorter response is sufficient.

\subsection{Generalization}

We assess the out-of-distribution generalization of the length reduction of \method with respect to STaR using the public test set of the MMLU-CF benchmark \citep[][see Section \ref{datasets}]{hendrycks2021measuringmassivemultitasklanguage}. As shown in Table \ref{table:mmlu}, \method achieves similar performance while generating 28\% to 36\% fewer tokens than STaR.

\begin{figure}[h]
\centering
\includegraphics[width=0.9\textwidth]{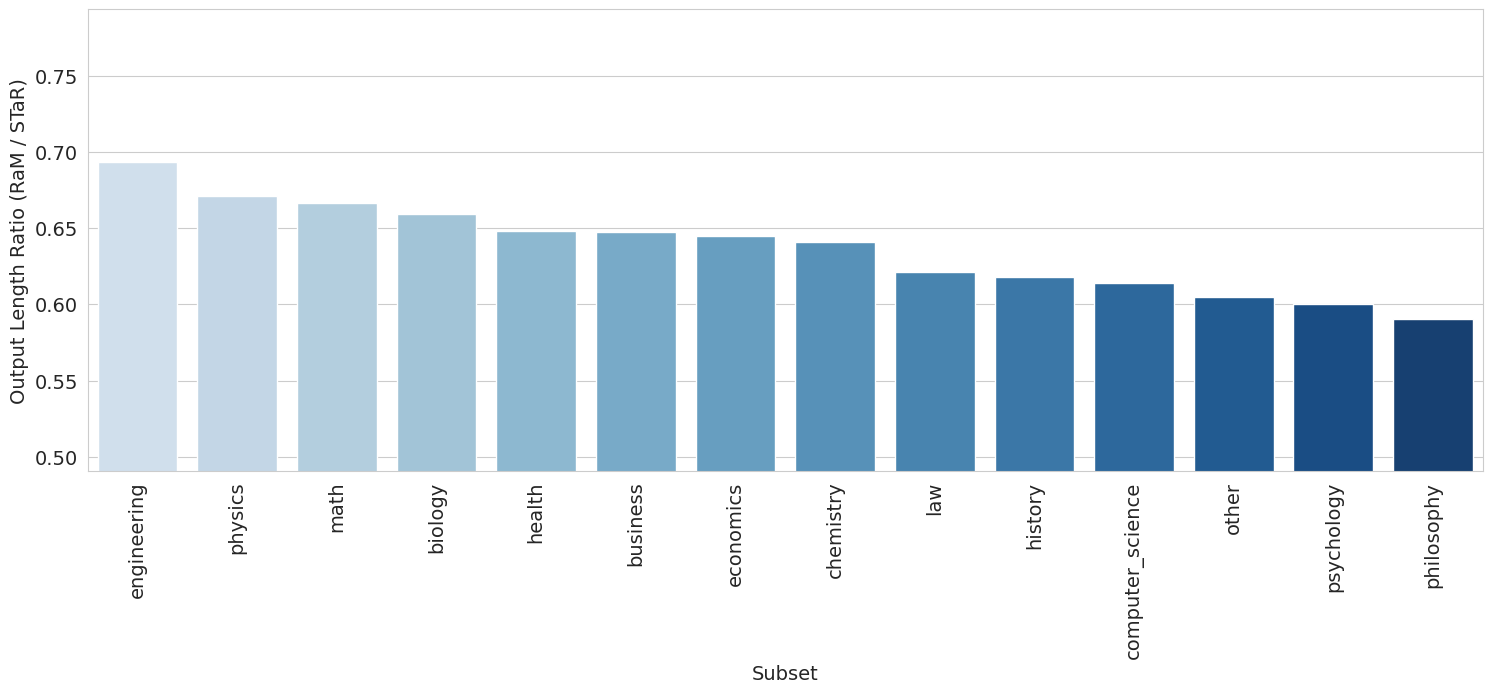}
\vspace{-1em}
\caption{{\bf Ratio of output length reduction.} \method’s output length relative to STaR’s, for Llama-3.2-3B. Tasks requiring heavier reasoning tend to exhibit a lower length reduction ratio, whereas those relying more on knowledge display a more pronounced length reduction.}
\label{mmlu}
\vspace{-1em}
\end{figure}

Looking at subdomains, Fig.~\ref{mmlu} shows the ratio between the output length of \method with respect to STaR. We can see that subjects that require multi-step reasoning (for example, physics or engineering) exhibit smaller reductions because they need detailed intermediate steps. In contrast, tasks focused primarily on recalling factual information tend to show a more pronounced length reduction.

\begin{table}
\centering

\begin{small}
\begin{tabular}{@{}clccc@{}}
\toprule
\textbf{Model}                   & \textbf{Method} & \textbf{Hard Split} & \textbf{Easy Split} & \textbf{Length Reduction (\%) $\uparrow$} \\ \hline
\multirow{3}{*}{Llama-3.2-3B}           
                                 & CoT Few-Shot   & 210.0 ($\pm$ 3.0)   & 170.0 ($\pm$ 3.0)   & 19.0 \\
                                  & STaR           & 195.0 ($\pm$ 5.0)   & 128.0 ($\pm$ 1.0)   & 34.4 \\
                                  & \method  & 145.0 ($\pm$ 3.0)   & 72.0  ($\pm$ 2.0)   & \textbf{50.3}          \\ \midrule
\multirow{1}{*}{Llama-3.2-3B-Instruct}           
                                 & CoT Prompting     & 222.0 ($\pm$ 4.0)   & 221.0 ($\pm$ 3.0)   & 0.5                   \\ \midrule
\multirow{3}{*}{Llama-3.1-8B} 
                                 & CoT Few-Shot   & 197.5 ($\pm$ 3.5)   & 151.0 ($\pm$ 1.0)   & 23.5 \\                        
                                  & STaR           & 202.0 ($\pm$ 5.0)   & 129.5 ($\pm$ 2.5)   & \textbf{35.9} \\
                                  & \method  & 139.0 ($\pm$ 3.0)   & 89.0  ($\pm$ 1.0)   & \textbf{36.0}  \\ \midrule
\multirow{1}{*}{Llama-3.1-8B-Instruct}           
                                 & CoT Prompting    & 228.0 ($\pm$ 6.0)   & 226.0 ($\pm$ 2.0)   & 0.9  \\ \bottomrule
\end{tabular}
\end{small}
\begin{small}
\caption{Length of generated reasoning chains across models and methods (mean number of tokens with 95\% confidence intervals). Adaptive methods should maximize the difference in length between the two distributions. \method reduces the overall length and increases the difference in the length distribution between the Hard and Easy splits (Length Reduction), demonstrating an improvement in the  ability to adapt reasoning length to task complexity.}
\label{table:comparison_length}
\end{small}
\end{table}

\begin{table}
\centering

\begin{small}
\begin{tabular}{@{}clccc@{}}
\toprule
\textbf{Model} & \textbf{Method} & \textbf{Accuracy (\%) $\uparrow$} & \textbf{\# Input Tokens $\downarrow$} & \textbf{\# Output Tokens $\downarrow$} \\ \midrule
\multirow{2}{*}{Llama-3.2-3B}
 & STaR & \textbf{27.0} ($\pm$ 2.4) & \textbf{75.1} ($\pm$ 0.9) & 150.6 ($\pm$ 4.1) \\
 & \method & \textbf{29.5} ($\pm$ 2.5) & \textbf{75.1} ($\pm$ 0.9) & \textbf{96.3} ($\pm$ 3.7) \\ \midrule
\multirow{2}{*}{Llama-3.1-8B}
 & STaR & \textbf{40.2} ($\pm$ 2.4) & \textbf{75.1} ($\pm$ 0.8) & 153.6 ($\pm$ 4.0) \\
 & \method & \textbf{36.4} ($\pm$ 2.4) & \textbf{75.1} ($\pm$ 0.9) & \textbf{110.6}  ($\pm$ 3.8) 
 \\ \bottomrule
\end{tabular}
\end{small}
\begin{small}
\caption{Comparison of STaR and \method based on rescaled accuracy and length metrics in an out-of-distribution setting on the MMLU-CF benchmark. We report the mean with 95\% confidence intervals. \method has similar performance but generates fewer output tokens.}
\end{small}
\label{table:mmlu}
\end{table}

%%%%%%%%%%%%%%%%%%%%%%%%%%%%%%%%%%%%%%%%%%%%%%%%%%%%%%%%%%%%
\section{Related Work}
%%%%%%%%%%%%%%%%%%%%%%%%%%%%%%%%%%%%%%%%%%%%%%%%%%%%%%%%%%%%

\subsection{Reducing inference costs}
The rising cost of deploying large language models (LLMs) has driven efforts to reduce inference costs. Techniques such as Speculative Decoding \citep{leviathan2023fastinferencetransformersspeculative} and Medusa \citep{cai2024medusasimplellminference} improve efficiency through parallelization, while Mixture of Experts \citep{6797059, zhou2022mixtureofexpertsexpertchoicerouting} activates only a subset of LLM parameters during decoding. Though effective, these methods require significant architectural changes and don't adapt computation based on task difficulty. Other approaches have developed neural architectures that enable adaptive computation \citep{graves2017adaptivecomputationtimerecurrent, banino2021pondernetlearningponder, dehghani2019universaltransformers, mohtashami2023cotformertokensattentionmake,schuster2022confidentadaptivelanguagemodeling} but involve new architectures or training methods. In contrast, our approach uses existing architectures and pretrained models, modifying only the fine-tuning process. More similar to our approach, model routing \citep{ong2024routellmlearningroutellms, jiang2024mixtralexperts} optimizes resource utilization based on query complexity by routing easier queries to smaller models and harder queries to larger ones. However, this necessitates multiple models and a router, while our approach trains a single model to adaptively adjust its own outputs to match task complexity. 

Finally, the concurrent “long-to-short’’ literature deliberately compresses reasoning paths while preserving accuracy. \citep{arora2025traininglanguagemodelsreason} shows that adding a length penalty to verifiable rewards can shorten the reasoning length of Large Reasoning Models in mathematical settings. Here, we focus on bootstrapping the reasoning process, teaching pretrained language models to reason efficiently from scratch by developing a reward that is agnostic to the answer format. CoT-Valve \citep{ma2025cotvalvelengthcompressiblechainofthoughttuning} and Claude 3.7 \citep{anthropicclaude} enable external control over the length of the reasoning chains (and therefore the computation budget), allowing a model to produce both long and short reasoning chains. Unlike these methods, our method enables the model to independently adapt the reasoning length according to the difficulty of the task. Concurrent work \citep{wu2025unlockingefficientlongtoshortllm, luo2025adar1hybridcotbileveladaptive} also shows that reasoning length compression can also be achieved by merging LLMs and Large Reasoning Models. Similarly, Synergy-of-Thoughts \citep{shang2024synergyofthoughtselicitingefficientreasoning} employed models of different sizes to adapt to the complexity of the task at hand. Our approach differs from all of the above on two fronts.  First, it keeps a single set of parameters: there is no external controller, merged checkpoint, or auxiliary model.  Second, adaptivity is learned through a principled Value-of-Computation reward, not by hard length constraints or heuristic escalation.

\subsection{Reasoning in LLMs}

Techniques such as Chain of Thought (CoT) and related methodologies  \citep{wei2023chainofthoughtpromptingelicitsreasoning, yao2023treethoughtsdeliberateproblem, madaan2023selfrefineiterativerefinementselffeedback, zheng2024stepbackevokingreasoning} have proven effective at enhancing LLM performance across a wide range of tasks. CoT boosts LLMs' performance on complex reasoning by guiding them through a series of intermediate reasoning steps, increasing inference costs to improve task performance. This method can be implemented through in-context learning \citep{wei2023chainofthoughtpromptingelicitsreasoning}, prompting \citep{kojima2023largelanguagemodelszeroshot}, or training \citep{li-etal-2023-symbolic}. The benefits of CoT can be attributed to both a greater computation depth \citep{goyal2024thinkspeaktraininglanguage, pfau2024letsthinkdotdot} and the semantic values of the thought tokens, which function as intermediate variables in the computation of the answer \citep{prystawski2023thinkstepstepreasoning}. However, recent studies have raised concerns regarding the meaningfulness of such reasoning chains in reaching the target solution, and whether models effectively utilize them to solve tasks \citep{turpin2023languagemodelsdontsay, paul2024makingreasoningmattermeasuring, sprague2024cotcotchainofthoughthelps}. We further demonstrate that standard prompting and training methods fail to teach the model to use CoT purposefully, resulting in inefficient inference. Moreover, reasoning in situations where it is unnecessary may also harm model performance \cite{liu2024mindstepbystep}.

Reasoning can also be used to bootstrap language models. Self-improving techniques \citep{huang2022largelanguagemodelsselfimprove, zelikman2022starbootstrappingreasoningreasoning} consist of generating reasoning chain-augmented answers for unlabeled questions and fine-tuning the LLM using those self-generated solutions as target outputs. Similar techniques, such as ReST \citep{gulcehre2023reinforcedselftrainingrestlanguage}, can also be used to better align LLMs with human preferences and needs. Our approach builds on these techniques to optimize the inference cost of reasoning  in addition to  task performance.

Another related work, Quiet-STaR \citep{zelikman2024quietstarlanguagemodelsteach}, has improved LLM reasoning performance by modifying pretraining to generate intermediate reasoning sequences between tokens. While effective for downstream tasks, it increases computational costs by generating reasoning chains at every step, even when unnecessary. More recently, chat models that ``think before answering'' have been developed \citep{o1systemcard, claude37sonnet, deepseekai2025deepseekr1incentivizingreasoningcapability}, using inference-time computation to enhance their outputs. Although these models outperform others, they expend more computational resources even when it may not be necessary. Our method could be incorporated into their training process to help the model determine when this additional computation is genuinely beneficial.

%%%%%%%%%%%%%%%%%%%%%%%%%%%%%%%%%%%%%%%%%%%%%%%%%%%%%%%%%%%%
\section{Limitations and Future Work}
%%%%%%%%%%%%%%%%%%%%%%%%%%%%%%%%%%%%%%%%%%%%%%%%%%%%%%%%%%%%

While our experiments show that rational metareasoning can successfully reduce inference costs in large language models, some limitations remain. First, our approach has only been tested on well-established datasets in science, commonsense reasoning, logic, and mathematics. Its effectiveness in other contexts has yet to be demonstrated. One particularly relevant example is the agentic setting, where LLMs act autonomously in complex digital environments \citep{yao2023reactsynergizingreasoningacting, schick2023toolformerlanguagemodelsteach}. Adapting our method to this context would require incorporating the cost of tool use (e.g., API calls) into the reward function to encourage the model to minimize unnecessary resource consumption. Second, while the VoC reward function is highly flexible and can be readily combined with various learning algorithms, our work focused on Expert Iteration \citep{havrilla2024teachinglargelanguagemodels} due to its sample efficiency for reasoning tasks. Extending the VoC reward framework to other training approaches like PPO or DPO remains an open research area. Evaluating whether VoC-based rewards maintain or even improve performance in these settings would further clarify the robustness and generalizability of our approach.

%%%%%%%%%%%%%%%%%%%%%%%%%%%%%%%%%%%%%%%%%%%%%%%%%%%%%%%%%%%%
\section{Conclusion}
%%%%%%%%%%%%%%%%%%%%%%%%%%%%%%%%%%%%%%%%%%%%%%%%%%%%%%%%%%%%
We introduced a cognitively inspired reward function grounded in rational metareasoning, aimed at optimizing LLM inference by balancing performance and computational cost. Empirically, we showed that this approach substantially reduces the number of generated tokens and input context length, while consistently improving task accuracy across a diverse range of benchmarks. Most excitingly, our work demonstrates how cognitively-inspired reward functions can endow LLMs with desirable inference-time properties, opening a broad avenue of future work. Given its flexibility, this method could be integrated into instruction tuning to potentially enhance performance, even in scenarios where verifying the correctness of answers is challenging. Since the utility measure within the reward function can be tailored to prioritize any desired, measurable property, this approach offers the potential to guide models toward achieving these enhanced qualities while still benefiting from the reduced computational costs.

\newpage

\bibliography{colm2025_conference}

\begin{thebibliography}{69}
\providecommand{\natexlab}[1]{#1}
\providecommand{\url}[1]{\texttt{#1}}
\expandafter\ifx\csname urlstyle\endcsname\relax
  \providecommand{\doi}[1]{doi: #1}\else
  \providecommand{\doi}{doi: \begingroup \urlstyle{rm}\Url}\fi

\bibitem[Anthony et~al.(2017)Anthony, Tian, and Barber]{anthony2017thinkingfastslowdeep}
Thomas Anthony, Zheng Tian, and David Barber.
\newblock Thinking fast and slow with deep learning and tree search, 2017.
\newblock URL \url{https://arxiv.org/abs/1705.08439}.

\bibitem[Anthropic(2025)]{anthropicclaude}
Anthropic.
\newblock Claude 3.7 sonnet and claude code”, 2025.
\newblock URL \url{https://www.anthropic.com/news/claude-3-7-sonnet}.

\bibitem[Anthropic et~al.(2025)]{claude37sonnet}
Anthropic et~al.
\newblock Claude 3.7 sonnet system card, 2025.
\newblock URL \url{https://assets.anthropic.com/m/785e231869ea8b3b/original/claude-3-7-sonnet-system-card.pdf}.

\bibitem[Arora \& Zanette(2025)Arora and Zanette]{arora2025traininglanguagemodelsreason}
Daman Arora and Andrea Zanette.
\newblock Training language models to reason efficiently, 2025.
\newblock URL \url{https://arxiv.org/abs/2502.04463}.

\bibitem[Banino et~al.(2021)Banino, Balaguer, and Blundell]{banino2021pondernetlearningponder}
Andrea Banino, Jan Balaguer, and Charles Blundell.
\newblock Pondernet: Learning to ponder, 2021.
\newblock URL \url{https://arxiv.org/abs/2107.05407}.

\bibitem[Cai et~al.(2024)Cai, Li, Geng, Peng, Lee, Chen, and Dao]{cai2024medusasimplellminference}
Tianle Cai, Yuhong Li, Zhengyang Geng, Hongwu Peng, Jason~D. Lee, Deming Chen, and Tri Dao.
\newblock Medusa: Simple llm inference acceleration framework with multiple decoding heads, 2024.
\newblock URL \url{https://arxiv.org/abs/2401.10774}.

\bibitem[Callaway et~al.(2018)Callaway, Gul, Krueger, Griffiths, and Lieder]{callaway2018learning}
Frederick Callaway, Sayan Gul, Paul Krueger, Thomas~L. Griffiths, and Falk Lieder.
\newblock Learning to select computations.
\newblock In \emph{34th Conference on Uncertainty in Artificial Intelligence (UAI 2018)}, pp.\  776--785. Curran Associates, Inc., 2018.

\bibitem[Callaway et~al.(2021)Callaway, Rangel, and Griffiths]{callaway2021fixation}
Frederick Callaway, Antonio Rangel, and Thomas~L Griffiths.
\newblock Fixation patterns in simple choice reflect optimal information sampling.
\newblock \emph{PLoS computational biology}, 17\penalty0 (3):\penalty0 e1008863, 2021.

\bibitem[Callaway et~al.(2022)Callaway, van Opheusden, Gul, Das, Krueger, Griffiths, and Lieder]{callaway2022rational}
Frederick Callaway, Bas van Opheusden, Sayan Gul, Priyam Das, Paul~M Krueger, Thomas~L Griffiths, and Falk Lieder.
\newblock Rational use of cognitive resources in human planning.
\newblock \emph{Nature Human Behaviour}, 6\penalty0 (8):\penalty0 1112--1125, 2022.

\bibitem[Chen et~al.(2024)Chen, Zhu, Soselia, Chen, Zhou, Goldstein, Huang, Shoeybi, and Catanzaro]{chen2024odindisentangledrewardmitigates}
Lichang Chen, Chen Zhu, Davit Soselia, Jiuhai Chen, Tianyi Zhou, Tom Goldstein, Heng Huang, Mohammad Shoeybi, and Bryan Catanzaro.
\newblock Odin: Disentangled reward mitigates hacking in rlhf, 2024.
\newblock URL \url{https://arxiv.org/abs/2402.07319}.

\bibitem[Chowdhery et~al.(2022)Chowdhery, Narang, Devlin, Bosma, Mishra, Roberts, Barham, Chung, Sutton, Gehrmann, et~al.]{chowdhery2022palmscalinglanguagemodeling}
Aakanksha Chowdhery, Sharan Narang, Jacob Devlin, Maarten Bosma, Gaurav Mishra, Adam Roberts, Paul Barham, Hyung~Won Chung, Charles Sutton, Sebastian Gehrmann, et~al.
\newblock Palm: Scaling language modeling with pathways, 2022.
\newblock URL \url{https://arxiv.org/abs/2204.02311}.

\bibitem[Clark et~al.(2018)Clark, Cowhey, Etzioni, Khot, Sabharwal, Schoenick, and Tafjord]{clark2018thinksolvedquestionanswering}
Peter Clark, Isaac Cowhey, Oren Etzioni, Tushar Khot, Ashish Sabharwal, Carissa Schoenick, and Oyvind Tafjord.
\newblock Think you have solved question answering? try arc, the ai2 reasoning challenge, 2018.
\newblock URL \url{https://arxiv.org/abs/1803.05457}.

\bibitem[Cobbe et~al.(2021)Cobbe, Kosaraju, Bavarian, Chen, Jun, Kaiser, Plappert, Tworek, Hilton, Nakano, Hesse, and Schulman]{cobbe2021trainingverifierssolvemath}
Karl Cobbe, Vineet Kosaraju, Mohammad Bavarian, Mark Chen, Heewoo Jun, Lukasz Kaiser, Matthias Plappert, Jerry Tworek, Jacob Hilton, Reiichiro Nakano, Christopher Hesse, and John Schulman.
\newblock Training verifiers to solve math word problems, 2021.
\newblock URL \url{https://arxiv.org/abs/2110.14168}.

\bibitem[{de Vries}(2023)]{DEVRIES20232191}
Alex {de Vries}.
\newblock The growing energy footprint of artificial intelligence.
\newblock \emph{Joule}, 7\penalty0 (10):\penalty0 2191--2194, 2023.
\newblock ISSN 2542-4351.
\newblock \doi{https://doi.org/10.1016/j.joule.2023.09.004}.
\newblock URL \url{https://www.sciencedirect.com/science/article/pii/S2542435123003653}.

\bibitem[DeepSeek-AI et~al.(2025)]{deepseekai2025deepseekr1incentivizingreasoningcapability}
DeepSeek-AI et~al.
\newblock Deepseek-r1: Incentivizing reasoning capability in llms via reinforcement learning, 2025.
\newblock URL \url{https://arxiv.org/abs/2501.12948}.

\bibitem[Dehghani et~al.(2019)Dehghani, Gouws, Vinyals, Uszkoreit, and Łukasz Kaiser]{dehghani2019universaltransformers}
Mostafa Dehghani, Stephan Gouws, Oriol Vinyals, Jakob Uszkoreit, and Łukasz Kaiser.
\newblock Universal transformers, 2019.
\newblock URL \url{https://arxiv.org/abs/1807.03819}.

\bibitem[Dubey et~al.(2024)Dubey, Jauhri, Pandey, Kadian, Al-Dahle, Letman, Mathur, Schelten, Yang, Fan, et~al.]{dubey2024llama3herdmodels}
Abhimanyu Dubey, Abhinav Jauhri, Abhinav Pandey, Abhishek Kadian, Ahmad Al-Dahle, Aiesha Letman, Akhil Mathur, Alan Schelten, Amy Yang, Angela Fan, et~al.
\newblock The llama 3 herd of models, 2024.
\newblock URL \url{https://arxiv.org/abs/2407.21783}.

\bibitem[Goyal et~al.(2024)Goyal, Ji, Rawat, Menon, Kumar, and Nagarajan]{goyal2024thinkspeaktraininglanguage}
Sachin Goyal, Ziwei Ji, Ankit~Singh Rawat, Aditya~Krishna Menon, Sanjiv Kumar, and Vaishnavh Nagarajan.
\newblock Think before you speak: Training language models with pause tokens, 2024.
\newblock URL \url{https://arxiv.org/abs/2310.02226}.

\bibitem[Graves(2017)]{graves2017adaptivecomputationtimerecurrent}
Alex Graves.
\newblock Adaptive computation time for recurrent neural networks, 2017.
\newblock URL \url{https://arxiv.org/abs/1603.08983}.

\bibitem[Griffiths(2020)]{Griffiths2020}
Thomas~L. Griffiths.
\newblock Understanding human intelligence through human limitations.
\newblock \emph{Trends in Cognitive Sciences}, 24\penalty0 (11):\penalty0 873–883, 2020.
\newblock ISSN 1364-6613.
\newblock \doi{10.1016/j.tics.2020.09.001}.
\newblock URL \url{http://dx.doi.org/10.1016/j.tics.2020.09.001}.

\bibitem[Griffiths et~al.(2019)Griffiths, Callaway, Chang, Grant, Krueger, and Lieder]{Griffiths2019}
Thomas~L Griffiths, Frederick Callaway, Michael~B Chang, Erin Grant, Paul~M Krueger, and Falk Lieder.
\newblock Doing more with less: meta-reasoning and meta-learning in humans and machines.
\newblock \emph{Current Opinion in Behavioral Sciences}, 29:\penalty0 24–30, 2019.
\newblock ISSN 2352-1546.
\newblock \doi{10.1016/j.cobeha.2019.01.005}.
\newblock URL \url{http://dx.doi.org/10.1016/j.cobeha.2019.01.005}.

\bibitem[Gulcehre et~al.(2023)Gulcehre, Paine, Srinivasan, Konyushkova, Weerts, Sharma, Siddhant, Ahern, Wang, Gu, Macherey, Doucet, Firat, and de~Freitas]{gulcehre2023reinforcedselftrainingrestlanguage}
Caglar Gulcehre, Tom~Le Paine, Srivatsan Srinivasan, Ksenia Konyushkova, Lotte Weerts, Abhishek Sharma, Aditya Siddhant, Alex Ahern, Miaosen Wang, Chenjie Gu, Wolfgang Macherey, Arnaud Doucet, Orhan Firat, and Nando de~Freitas.
\newblock Reinforced self-training (rest) for language modeling, 2023.
\newblock URL \url{https://arxiv.org/abs/2308.08998}.

\bibitem[Havrilla et~al.(2024)Havrilla, Du, Raparthy, Nalmpantis, Dwivedi-Yu, Zhuravinskyi, Hambro, Sukhbaatar, and Raileanu]{havrilla2024teachinglargelanguagemodels}
Alex Havrilla, Yuqing Du, Sharath~Chandra Raparthy, Christoforos Nalmpantis, Jane Dwivedi-Yu, Maksym Zhuravinskyi, Eric Hambro, Sainbayar Sukhbaatar, and Roberta Raileanu.
\newblock Teaching large language models to reason with reinforcement learning, 2024.
\newblock URL \url{https://arxiv.org/abs/2403.04642}.

\bibitem[Hendrycks et~al.(2021)Hendrycks, Burns, Basart, Zou, Mazeika, Song, and Steinhardt]{hendrycks2021measuringmassivemultitasklanguage}
Dan Hendrycks, Collin Burns, Steven Basart, Andy Zou, Mantas Mazeika, Dawn Song, and Jacob Steinhardt.
\newblock Measuring massive multitask language understanding, 2021.
\newblock URL \url{https://arxiv.org/abs/2009.03300}.

\bibitem[Huang et~al.(2022)Huang, Gu, Hou, Wu, Wang, Yu, and Han]{huang2022largelanguagemodelsselfimprove}
Jiaxin Huang, Shixiang~Shane Gu, Le~Hou, Yuexin Wu, Xuezhi Wang, Hongkun Yu, and Jiawei Han.
\newblock Large language models can self-improve, 2022.
\newblock URL \url{https://arxiv.org/abs/2210.11610}.

\bibitem[Jacobs et~al.(1991)Jacobs, Jordan, Nowlan, and Hinton]{6797059}
Robert~A. Jacobs, Michael~I. Jordan, Steven~J. Nowlan, and Geoffrey~E. Hinton.
\newblock Adaptive mixtures of local experts.
\newblock \emph{Neural Computation}, 3\penalty0 (1):\penalty0 79--87, 1991.
\newblock \doi{10.1162/neco.1991.3.1.79}.

\bibitem[Jiang et~al.(2024)Jiang, Sablayrolles, Roux, Mensch, Savary, Bamford, Chaplot, de~las Casas, Hanna, Bressand, Lengyel, Bour, Lample, Lavaud, Saulnier, Lachaux, Stock, Subramanian, Yang, Antoniak, Scao, Gervet, Lavril, Wang, Lacroix, and Sayed]{jiang2024mixtralexperts}
Albert~Q. Jiang, Alexandre Sablayrolles, Antoine Roux, Arthur Mensch, Blanche Savary, Chris Bamford, Devendra~Singh Chaplot, Diego de~las Casas, Emma~Bou Hanna, Florian Bressand, Gianna Lengyel, Guillaume Bour, Guillaume Lample, Lélio~Renard Lavaud, Lucile Saulnier, Marie-Anne Lachaux, Pierre Stock, Sandeep Subramanian, Sophia Yang, Szymon Antoniak, Teven~Le Scao, Théophile Gervet, Thibaut Lavril, Thomas Wang, Timothée Lacroix, and William~El Sayed.
\newblock Mixtral of experts, 2024.
\newblock URL \url{https://arxiv.org/abs/2401.04088}.

\bibitem[Kahneman(2014)]{Kahneman}
Daniel Kahneman.
\newblock Thinking, fast and slow.
\newblock \emph{Stat. Pap. (Berl)}, 55\penalty0 (3):\penalty0 915--915, 2014.

\bibitem[Kojima et~al.(2023)Kojima, Gu, Reid, Matsuo, and Iwasawa]{kojima2023largelanguagemodelszeroshot}
Takeshi Kojima, Shixiang~Shane Gu, Machel Reid, Yutaka Matsuo, and Yusuke Iwasawa.
\newblock Large language models are zero-shot reasoners, 2023.
\newblock URL \url{https://arxiv.org/abs/2205.11916}.

\bibitem[Leviathan et~al.(2023)Leviathan, Kalman, and Matias]{leviathan2023fastinferencetransformersspeculative}
Yaniv Leviathan, Matan Kalman, and Yossi Matias.
\newblock Fast inference from transformers via speculative decoding, 2023.
\newblock URL \url{https://arxiv.org/abs/2211.17192}.

\bibitem[Li et~al.(2023)Li, Hessel, Yu, Ren, Chang, and Choi]{li-etal-2023-symbolic}
Liunian~Harold Li, Jack Hessel, Youngjae Yu, Xiang Ren, Kai-Wei Chang, and Yejin Choi.
\newblock Symbolic chain-of-thought distillation: Small models can also {``}think{''} step-by-step.
\newblock In Anna Rogers, Jordan Boyd-Graber, and Naoaki Okazaki (eds.), \emph{Proceedings of the 61st Annual Meeting of the Association for Computational Linguistics (Volume 1: Long Papers)}, pp.\  2665--2679, Toronto, Canada, 2023. Association for Computational Linguistics.
\newblock \doi{10.18653/v1/2023.acl-long.150}.
\newblock URL \url{https://aclanthology.org/2023.acl-long.150}.

\bibitem[Lieder \& Griffiths(2017)Lieder and Griffiths]{Lieder2017}
Falk Lieder and Thomas~L. Griffiths.
\newblock Strategy selection as rational metareasoning.
\newblock \emph{Psychological Review}, 124\penalty0 (6):\penalty0 762–794, 2017.
\newblock ISSN 0033-295X.
\newblock \doi{10.1037/rev0000075}.
\newblock URL \url{http://dx.doi.org/10.1037/rev0000075}.

\bibitem[Lieder et~al.(2018)Lieder, Shenhav, Musslick, and Griffiths]{Lieder2018}
Falk Lieder, Amitai Shenhav, Sebastian Musslick, and Thomas~L. Griffiths.
\newblock Rational metareasoning and the plasticity of cognitive control.
\newblock \emph{PLOS Computational Biology}, 14\penalty0 (4):\penalty0 e1006043, 2018.
\newblock ISSN 1553-7358.
\newblock \doi{10.1371/journal.pcbi.1006043}.
\newblock URL \url{http://dx.doi.org/10.1371/journal.pcbi.1006043}.

\bibitem[Liu et~al.(2024)Liu, Geng, Wu, Sucholutsky, Lombrozo, and Griffiths]{liu2024mindstepbystep}
Ryan Liu, Jiayi Geng, Addison~J. Wu, Ilia Sucholutsky, Tania Lombrozo, and Thomas~L. Griffiths.
\newblock Mind your step (by step): Chain-of-thought can reduce performance on tasks where thinking makes humans worse, 2024.
\newblock URL \url{https://arxiv.org/abs/2410.21333}.

\bibitem[Luo et~al.(2025)Luo, He, Wang, Yang, Liu, Tan, Cao, Tao, and Shen]{luo2025adar1hybridcotbileveladaptive}
Haotian Luo, Haiying He, Yibo Wang, Jinluan Yang, Rui Liu, Naiqiang Tan, Xiaochun Cao, Dacheng Tao, and Li~Shen.
\newblock Ada-r1: Hybrid-cot via bi-level adaptive reasoning optimization, 2025.
\newblock URL \url{https://arxiv.org/abs/2504.21659}.

\bibitem[Ma et~al.(2025)Ma, Wan, Yu, Fang, and Wang]{ma2025cotvalvelengthcompressiblechainofthoughttuning}
Xinyin Ma, Guangnian Wan, Runpeng Yu, Gongfan Fang, and Xinchao Wang.
\newblock Cot-valve: Length-compressible chain-of-thought tuning, 2025.
\newblock URL \url{https://arxiv.org/abs/2502.09601}.

\bibitem[Madaan et~al.(2023)Madaan, Tandon, Gupta, Hallinan, Gao, Wiegreffe, Alon, Dziri, Prabhumoye, Yang, Gupta, Majumder, Hermann, Welleck, Yazdanbakhsh, and Clark]{madaan2023selfrefineiterativerefinementselffeedback}
Aman Madaan, Niket Tandon, Prakhar Gupta, Skyler Hallinan, Luyu Gao, Sarah Wiegreffe, Uri Alon, Nouha Dziri, Shrimai Prabhumoye, Yiming Yang, Shashank Gupta, Bodhisattwa~Prasad Majumder, Katherine Hermann, Sean Welleck, Amir Yazdanbakhsh, and Peter Clark.
\newblock Self-refine: Iterative refinement with self-feedback, 2023.
\newblock URL \url{https://arxiv.org/abs/2303.17651}.

\bibitem[Mohtashami et~al.(2023)Mohtashami, Pagliardini, and Jaggi]{mohtashami2023cotformertokensattentionmake}
Amirkeivan Mohtashami, Matteo Pagliardini, and Martin Jaggi.
\newblock Cotformer: More tokens with attention make up for less depth, 2023.
\newblock URL \url{https://arxiv.org/abs/2310.10845}.

\bibitem[Ong et~al.(2024)Ong, Almahairi, Wu, Chiang, Wu, Gonzalez, Kadous, and Stoica]{ong2024routellmlearningroutellms}
Isaac Ong, Amjad Almahairi, Vincent Wu, Wei-Lin Chiang, Tianhao Wu, Joseph~E. Gonzalez, M~Waleed Kadous, and Ion Stoica.
\newblock Routellm: Learning to route llms with preference data, 2024.
\newblock URL \url{https://arxiv.org/abs/2406.18665}.

\bibitem[OpenAI(2024)]{o1systemcard}
OpenAI.
\newblock Openai o1 system card.
\newblock \emph{preprint}, 2024.
\newblock URL \url{https://openai.com/index/openai-o1-system-card/}.

\bibitem[OpenAI et~al.(2024)OpenAI, Achiam, Adler, Agarwal, Ahmad, Akkaya, Aleman, Almeida, Altenschmidt, Altman, et~al.]{openai2024gpt4technicalreport}
OpenAI, Josh Achiam, Steven Adler, Sandhini Agarwal, Lama Ahmad, Ilge Akkaya, Florencia~Leoni Aleman, Diogo Almeida, Janko Altenschmidt, Sam Altman, et~al.
\newblock Gpt-4 technical report, 2024.
\newblock URL \url{https://arxiv.org/abs/2303.08774}.

\bibitem[Park et~al.(2024)Park, Rafailov, Ermon, and Finn]{park2024disentanglinglengthqualitydirect}
Ryan Park, Rafael Rafailov, Stefano Ermon, and Chelsea Finn.
\newblock Disentangling length from quality in direct preference optimization, 2024.
\newblock URL \url{https://arxiv.org/abs/2403.19159}.

\bibitem[Paul et~al.(2024)Paul, West, Bosselut, and Faltings]{paul2024makingreasoningmattermeasuring}
Debjit Paul, Robert West, Antoine Bosselut, and Boi Faltings.
\newblock Making reasoning matter: Measuring and improving faithfulness of chain-of-thought reasoning, 2024.
\newblock URL \url{https://arxiv.org/abs/2402.13950}.

\bibitem[Pfau et~al.(2024)Pfau, Merrill, and Bowman]{pfau2024letsthinkdotdot}
Jacob Pfau, William Merrill, and Samuel~R. Bowman.
\newblock Let's think dot by dot: Hidden computation in transformer language models, 2024.
\newblock URL \url{https://arxiv.org/abs/2404.15758}.

\bibitem[Prystawski et~al.(2023)Prystawski, Li, and Goodman]{prystawski2023thinkstepstepreasoning}
Ben Prystawski, Michael~Y. Li, and Noah~D. Goodman.
\newblock Why think step by step? reasoning emerges from the locality of experience, 2023.
\newblock URL \url{https://arxiv.org/abs/2304.03843}.

\bibitem[Russek et~al.(2022)Russek, Acosta-Kane, van Opheusden, Mattar, and Griffiths]{russek2022time}
Evan Russek, Daniel Acosta-Kane, Bas van Opheusden, Marcelo~G Mattar, and Tom Griffiths.
\newblock Time spent thinking in online chess reflects the value of computation.
\newblock 2022.

\bibitem[Russell \& Wefald(1991)Russell and Wefald]{Russell1991}
Stuart Russell and Eric Wefald.
\newblock Principles of metareasoning.
\newblock \emph{Artificial Intelligence}, 49\penalty0 (1–3):\penalty0 361–395, 1991.
\newblock ISSN 0004-3702.
\newblock \doi{10.1016/0004-3702(91)90015-c}.
\newblock URL \url{http://dx.doi.org/10.1016/0004-3702(91)90015-C}.

\bibitem[Russell(1997)]{RUSSELL1997}
Stuart~J. Russell.
\newblock Rationality and intelligence.
\newblock \emph{Artificial Intelligence}, 94\penalty0 (1–2):\penalty0 57–77, 1997.
\newblock ISSN 0004-3702.
\newblock \doi{10.1016/s0004-3702(97)00026-x}.
\newblock URL \url{http://dx.doi.org/10.1016/S0004-3702(97)00026-X}.

\bibitem[Schick et~al.(2023)Schick, Dwivedi-Yu, Dessì, Raileanu, Lomeli, Zettlemoyer, Cancedda, and Scialom]{schick2023toolformerlanguagemodelsteach}
Timo Schick, Jane Dwivedi-Yu, Roberto Dessì, Roberta Raileanu, Maria Lomeli, Luke Zettlemoyer, Nicola Cancedda, and Thomas Scialom.
\newblock Toolformer: Language models can teach themselves to use tools, 2023.
\newblock URL \url{https://arxiv.org/abs/2302.04761}.

\bibitem[Schuster et~al.(2022)Schuster, Fisch, Gupta, Dehghani, Bahri, Tran, Tay, and Metzler]{schuster2022confidentadaptivelanguagemodeling}
Tal Schuster, Adam Fisch, Jai Gupta, Mostafa Dehghani, Dara Bahri, Vinh~Q. Tran, Yi~Tay, and Donald Metzler.
\newblock Confident adaptive language modeling, 2022.
\newblock URL \url{https://arxiv.org/abs/2207.07061}.

\bibitem[Shang et~al.(2024)Shang, Li, Xu, and Li]{shang2024synergyofthoughtselicitingefficientreasoning}
Yu~Shang, Yu~Li, Fengli Xu, and Yong Li.
\newblock Synergy-of-thoughts: Eliciting efficient reasoning in hybrid language models, 2024.
\newblock URL \url{https://arxiv.org/abs/2402.02563}.

\bibitem[Singhal et~al.(2024)Singhal, Goyal, Xu, and Durrett]{singhal2024longwaygoinvestigating}
Prasann Singhal, Tanya Goyal, Jiacheng Xu, and Greg Durrett.
\newblock A long way to go: Investigating length correlations in rlhf, 2024.
\newblock URL \url{https://arxiv.org/abs/2310.03716}.

\bibitem[Snell et~al.(2024)Snell, Lee, Xu, and Kumar]{snell2024scalingllmtesttimecompute}
Charlie Snell, Jaehoon Lee, Kelvin Xu, and Aviral Kumar.
\newblock Scaling llm test-time compute optimally can be more effective than scaling model parameters, 2024.
\newblock URL \url{https://arxiv.org/abs/2408.03314}.

\bibitem[Sprague et~al.(2024)Sprague, Yin, Rodriguez, Jiang, Wadhwa, Singhal, Zhao, Ye, Mahowald, and Durrett]{sprague2024cotcotchainofthoughthelps}
Zayne Sprague, Fangcong Yin, Juan~Diego Rodriguez, Dongwei Jiang, Manya Wadhwa, Prasann Singhal, Xinyu Zhao, Xi~Ye, Kyle Mahowald, and Greg Durrett.
\newblock To cot or not to cot? chain-of-thought helps mainly on math and symbolic reasoning, 2024.
\newblock URL \url{https://arxiv.org/abs/2409.12183}.

\bibitem[Tafjord et~al.(2021)Tafjord, Mishra, and Clark]{tafjord2021proofwritergeneratingimplicationsproofs}
Oyvind Tafjord, Bhavana~Dalvi Mishra, and Peter Clark.
\newblock Proofwriter: Generating implications, proofs, and abductive statements over natural language, 2021.
\newblock URL \url{https://arxiv.org/abs/2012.13048}.

\bibitem[Talmor et~al.(2019)Talmor, Herzig, Lourie, and Berant]{talmor2019commonsenseqaquestionansweringchallenge}
Alon Talmor, Jonathan Herzig, Nicholas Lourie, and Jonathan Berant.
\newblock Commonsenseqa: A question answering challenge targeting commonsense knowledge, 2019.
\newblock URL \url{https://arxiv.org/abs/1811.00937}.

\bibitem[Tang et~al.(2024)Tang, Guo, Zheng, Calandriello, Cao, Tarassov, Munos, Ávila Pires, Valko, Cheng, and Dabney]{tang2024understandingperformancegaponline}
Yunhao Tang, Daniel~Zhaohan Guo, Zeyu Zheng, Daniele Calandriello, Yuan Cao, Eugene Tarassov, Rémi Munos, Bernardo Ávila Pires, Michal Valko, Yong Cheng, and Will Dabney.
\newblock Understanding the performance gap between online and offline alignment algorithms, 2024.
\newblock URL \url{https://arxiv.org/abs/2405.08448}.

\bibitem[Turpin et~al.(2023)Turpin, Michael, Perez, and Bowman]{turpin2023languagemodelsdontsay}
Miles Turpin, Julian Michael, Ethan Perez, and Samuel~R. Bowman.
\newblock Language models don't always say what they think: Unfaithful explanations in chain-of-thought prompting, 2023.
\newblock URL \url{https://arxiv.org/abs/2305.04388}.

\bibitem[Verdecchia et~al.(2023)Verdecchia, Sallou, and Cruz]{Verdecchia2023}
Roberto Verdecchia, June Sallou, and Luís Cruz.
\newblock A systematic review of green <scp>ai</scp>.
\newblock \emph{WIREs Data Mining and Knowledge Discovery}, 13\penalty0 (4), 2023.
\newblock ISSN 1942-4795.
\newblock \doi{10.1002/widm.1507}.
\newblock URL \url{http://dx.doi.org/10.1002/widm.1507}.

\bibitem[Wan et~al.(2024)Wan, Wang, Liu, Alam, Zheng, Liu, Qu, Yan, Zhu, Zhang, Chowdhury, and Zhang]{wan2024efficientlargelanguagemodels}
Zhongwei Wan, Xin Wang, Che Liu, Samiul Alam, Yu~Zheng, Jiachen Liu, Zhongnan Qu, Shen Yan, Yi~Zhu, Quanlu Zhang, Mosharaf Chowdhury, and Mi~Zhang.
\newblock Efficient large language models: A survey, 2024.
\newblock URL \url{https://arxiv.org/abs/2312.03863}.

\bibitem[Wei et~al.(2023)Wei, Wang, Schuurmans, Bosma, Ichter, Xia, Chi, Le, and Zhou]{wei2023chainofthoughtpromptingelicitsreasoning}
Jason Wei, Xuezhi Wang, Dale Schuurmans, Maarten Bosma, Brian Ichter, Fei Xia, Ed~Chi, Quoc Le, and Denny Zhou.
\newblock Chain-of-thought prompting elicits reasoning in large language models, 2023.
\newblock URL \url{https://arxiv.org/abs/2201.11903}.

\bibitem[Wu et~al.(2025)Wu, Yao, Liu, Liu, Fu, Han, Li, Zhen, Zhong, and Yuan]{wu2025unlockingefficientlongtoshortllm}
Han Wu, Yuxuan Yao, Shuqi Liu, Zehua Liu, Xiaojin Fu, Xiongwei Han, Xing Li, Hui-Ling Zhen, Tao Zhong, and Mingxuan Yuan.
\newblock Unlocking efficient long-to-short llm reasoning with model merging, 2025.
\newblock URL \url{https://arxiv.org/abs/2503.20641}.

\bibitem[Yao et~al.(2023{\natexlab{a}})Yao, Yu, Zhao, Shafran, Griffiths, Cao, and Narasimhan]{yao2023treethoughtsdeliberateproblem}
Shunyu Yao, Dian Yu, Jeffrey Zhao, Izhak Shafran, Thomas~L. Griffiths, Yuan Cao, and Karthik Narasimhan.
\newblock Tree of thoughts: Deliberate problem solving with large language models, 2023{\natexlab{a}}.
\newblock URL \url{https://arxiv.org/abs/2305.10601}.

\bibitem[Yao et~al.(2023{\natexlab{b}})Yao, Zhao, Yu, Du, Shafran, Narasimhan, and Cao]{yao2023reactsynergizingreasoningacting}
Shunyu Yao, Jeffrey Zhao, Dian Yu, Nan Du, Izhak Shafran, Karthik Narasimhan, and Yuan Cao.
\newblock React: Synergizing reasoning and acting in language models, 2023{\natexlab{b}}.
\newblock URL \url{https://arxiv.org/abs/2210.03629}.

\bibitem[Zelikman et~al.(2022)Zelikman, Wu, Mu, and Goodman]{zelikman2022starbootstrappingreasoningreasoning}
Eric Zelikman, Yuhuai Wu, Jesse Mu, and Noah~D. Goodman.
\newblock Star: Bootstrapping reasoning with reasoning, 2022.
\newblock URL \url{https://arxiv.org/abs/2203.14465}.

\bibitem[Zelikman et~al.(2024)Zelikman, Harik, Shao, Jayasiri, Haber, and Goodman]{zelikman2024quietstarlanguagemodelsteach}
Eric Zelikman, Georges Harik, Yijia Shao, Varuna Jayasiri, Nick Haber, and Noah~D. Goodman.
\newblock Quiet-star: Language models can teach themselves to think before speaking, 2024.
\newblock URL \url{https://arxiv.org/abs/2403.09629}.

\bibitem[Zhao et~al.(2024)Zhao, Huang, Lv, Cui, Sun, Mao, Zhang, Xin, Yin, Li, and Wei]{zhao2024mmlucfcontaminationfreemultitasklanguage}
Qihao Zhao, Yangyu Huang, Tengchao Lv, Lei Cui, Qinzheng Sun, Shaoguang Mao, Xin Zhang, Ying Xin, Qiufeng Yin, Scarlett Li, and Furu Wei.
\newblock Mmlu-cf: A contamination-free multi-task language understanding benchmark, 2024.
\newblock URL \url{https://arxiv.org/abs/2412.15194}.

\bibitem[Zheng et~al.(2024)Zheng, Mishra, Chen, Cheng, Chi, Le, and Zhou]{zheng2024stepbackevokingreasoning}
Huaixiu~Steven Zheng, Swaroop Mishra, Xinyun Chen, Heng-Tze Cheng, Ed~H. Chi, Quoc~V Le, and Denny Zhou.
\newblock Take a step back: Evoking reasoning via abstraction in large language models, 2024.
\newblock URL \url{https://arxiv.org/abs/2310.06117}.

\bibitem[Zhou et~al.(2022)Zhou, Lei, Liu, Du, Huang, Zhao, Dai, Chen, Le, and Laudon]{zhou2022mixtureofexpertsexpertchoicerouting}
Yanqi Zhou, Tao Lei, Hanxiao Liu, Nan Du, Yanping Huang, Vincent Zhao, Andrew Dai, Zhifeng Chen, Quoc Le, and James Laudon.
\newblock Mixture-of-experts with expert choice routing, 2022.
\newblock URL \url{https://arxiv.org/abs/2202.09368}.

\end{thebibliography}
\bibliographystyle{colm2025_conference}

\appendix

\section{Complete Results Tables}

In this section, we present the comprehensive results tables.

\setlength{\tabcolsep}{2pt}

\begin{table}[H]
\begin{small}
\begin{tabular}{llcccc}
\toprule
\textbf{Model}                & \textbf{Method}      & \textbf{ARC}                & \textbf{Commonsenseqa}       & \textbf{GSM8K}              & \textbf{Proofwriter}       \\ \toprule
\multirow{5}{*}{Llama-3.2-3B}        & Direct Few Shot      & 70.1 ($\pm$ 1.8)            & 56.2 ($\pm$ 3.2)             & 7.6 ($\pm$ 1.6)             & 0.0 ($\pm$ 1.3)            \\
                              & Cot Few Shot         & 63.8 ($\pm$ 1.9)            & 43.3 ($\pm$ 3.3)             & 27.7 ($\pm$ 2.4)            & 4.2 ($\pm$ 4.5)            \\
                              & Direct SFT         & 82.8 ($\pm$ 1.2)            & 66.0 ($\pm$ 2.6)             & 41.2 ($\pm$ 2.8)            & 36.3 ($\pm$ 3.1)           \\
                              & STaR                 & 66.6 ($\pm$ 1.9)            & 52.5 ($\pm$ 3.5)             & 26.8 ($\pm$ 2.5)            & 43.6 ($\pm$ 4.5)           \\
                              & \method        & 71.2 ($\pm$ 1.7)            & 55.8 ($\pm$ 3.3)             & 32.3 ($\pm$ 2.5)            & 41.3 ($\pm$ 4.4)           \\ \midrule
\multirow{1}{*}{Llama-3.2-3B-Instruct}        & CoT Prompting        & 75.9 ($\pm$ 1.7)            & 58.1 ($\pm$ 3.2)             & 71.6 ($\pm$ 2.2)            & 7.3 ($\pm$ 4.4)            \\ \midrule
\multirow{5}{*}{Llama-3.1-8B} & Direct Few Shot      & 82.6 ($\pm$ 1.5)            & 66.4 ($\pm$ 3.0)             & 14.9 ($\pm$ 2.0)            & 19.4 ($\pm$ 4.4)           \\
                              & Cot Few Shot         & 80.9 ($\pm$ 1.5)            & 60.4 ($\pm$ 3.2)             & 52.8 ($\pm$ 2.6)            & 32.9 ($\pm$ 4.2)           \\
                              & STaR                 & 82.0 ($\pm$ 1.6)            & 64.0 ($\pm$ 3.0)             & 45.9 ($\pm$ 2.8)            & 55.8 ($\pm$ 4.1)           \\
                              & Metareasoning        & 80.3 ($\pm$ 1.5)            & 67.2 ($\pm$ 3.1)             & 53.7 ($\pm$ 2.5)            & 52.7 ($\pm$ 4.0)           \\ \midrule
\multirow{1}{*}{Llama-3.1-8B-Instruct}        & CoT Prompting        & 85.4 ($\pm$ 1.4)            & 66.6 ($\pm$ 3.0)             & 80.5 ($\pm$ 2.0)            & 11.2 ($\pm$ 4.3)           \\ 
                               \bottomrule

\end{tabular}
\end{small}
\caption{Comparison of accuracy of different methods across datasets and models. We report the mean with 95\% confidence scores.}
\label{table:accuracies}
\end{table}

\begin{table}[H]
\begin{small}
\begin{tabular}{llcccc}
\toprule
\textbf{Model}                & \textbf{Method}      & \textbf{ARC}           & \textbf{Commonsenseqa} & \textbf{GSM8K}        & \textbf{Proofwriter}      \\ \toprule
\multirow{5}{*}{Llama-3.2-3B}        
                                & Direct Few Shot      & 0.0 ($\pm$ 0.0)            & 0.0 ($\pm$ 0.0)              & 0.0 ($\pm$ 0.0)             & 0.0 ($\pm$ 0.0)            \\
                              & Cot Few Shot         & 169.6 ($\pm$ 2.3)          & 182.0 ($\pm$ 9.7)            & 236.9 ($\pm$ 8.0)           & 459.2 ($\pm$ 14.1)         \\
                              & STaR                 & 121.0 ($\pm$ 2.2)          & 130.0 ($\pm$ 1.4)            & 259.7 ($\pm$ 9.0)           & 380.0 ($\pm$ 7.2)          \\
                              & \method        & 77.4 ($\pm$ 1.5)           & 91.3 ($\pm$ 2.6)             & 184.1 ($\pm$ 7.8)           & 330.4 ($\pm$ 7.4)          \\ \midrule
\multirow{1}{*}{Llama-3.2-3B-Instruct}        
                                & CoT Prompting        & 230.8 ($\pm$ 2.2)          & 189.1 ($\pm$ 3.7)            & 192.1 ($\pm$ 4.6)           & 424.6 ($\pm$ 11.0)         \\ \midrule
\multirow{4}{*}{Meta-Llama-3-8B} 
                                & Direct Few Shot      & 0.0 ($\pm$ 0.0)            & 0.0 ($\pm$ 0.0)              & 0.0 ($\pm$ 0.0)             & 0.0 ($\pm$ 0.0)            \\
                              & Cot Few Shot         & 147.9 ($\pm$ 1.5)          & 153.6 ($\pm$ 6.8)            & 232.7 ($\pm$ 8.6)           & 472.1 ($\pm$ 16.8)         \\
                              & STaR                 & 119.6 ($\pm$ 2.0)          & 143.6 ($\pm$ 2.3)            & 256.9 ($\pm$ 7.5)           & 340.4 ($\pm$ 9.6)          \\
                              & Metareasoning        & 91.8 ($\pm$ 1.4)           & 93.1 ($\pm$ 3.2)             & 171.4 ($\pm$ 5.4)           & 228.7 ($\pm$ 10.1)         \\ \midrule
\multirow{1}{*}{Llama-3.1-8B-Instruct}        
                                & CoT Prompting        & 233.0 ($\pm$ 2.2)          & 200.7 ($\pm$ 4.3)            & 210.8 ($\pm$ 5.0)           & 435.1 ($\pm$ 11.7)         \\ \bottomrule
\end{tabular}
\end{small}
\caption{Comparison of mean output length of different methods across datasets and models. We report the mean with 95\% confidence scores.}
\label{table:output_lengths}
\end{table}

\section{CoT Prompting}\label{cot_prompting}
For the ``Instruct'' versions of our models we use the following prompt:

\texttt{\footnotesize{Answer the following question, thinking step by step to get to the answer. You can think however long you need, but answer as soon as you're ready. Use the minimum number of steps to get to the answer. Once you're finished thinking, you must end your response with 'The answer is [X]', where [X] is the final answer to the question.}}

\section{Qualitative examples}\label{qualitative_examples}

Below we show one example of question, reasoning chain and answer per dataset, generated with each method: CoT Few Shots (blue), STaR (yellow) and \method (green). The following examples were generated by Llama-3.2-3B \citep{dubey2024llama3herdmodels}. The reasoning chains shown here serve as intermediate computations that guide the model to its final answers. While not intended as direct explanations, the examples remain interpretable and demonstrate the effectiveness of the respective methods.

\begin{figure}
    \centering
    \includegraphics[width=\linewidth]{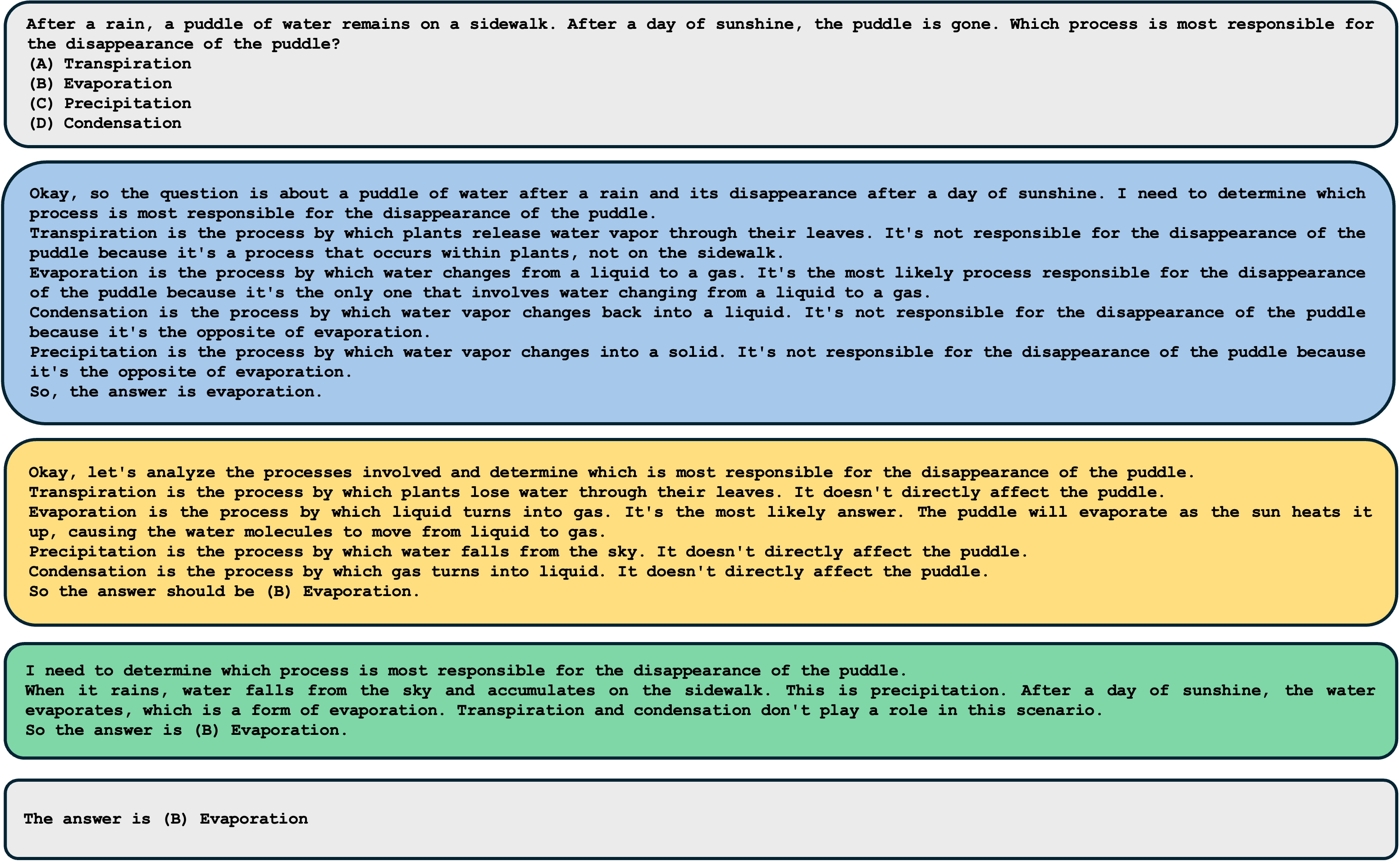} 
    \caption{Qualitative examples of reasoning processes for ARC dataset}
    \label{fig:examples_arc}
\end{figure}

\begin{figure}
    \centering

    \includegraphics[width=\linewidth]{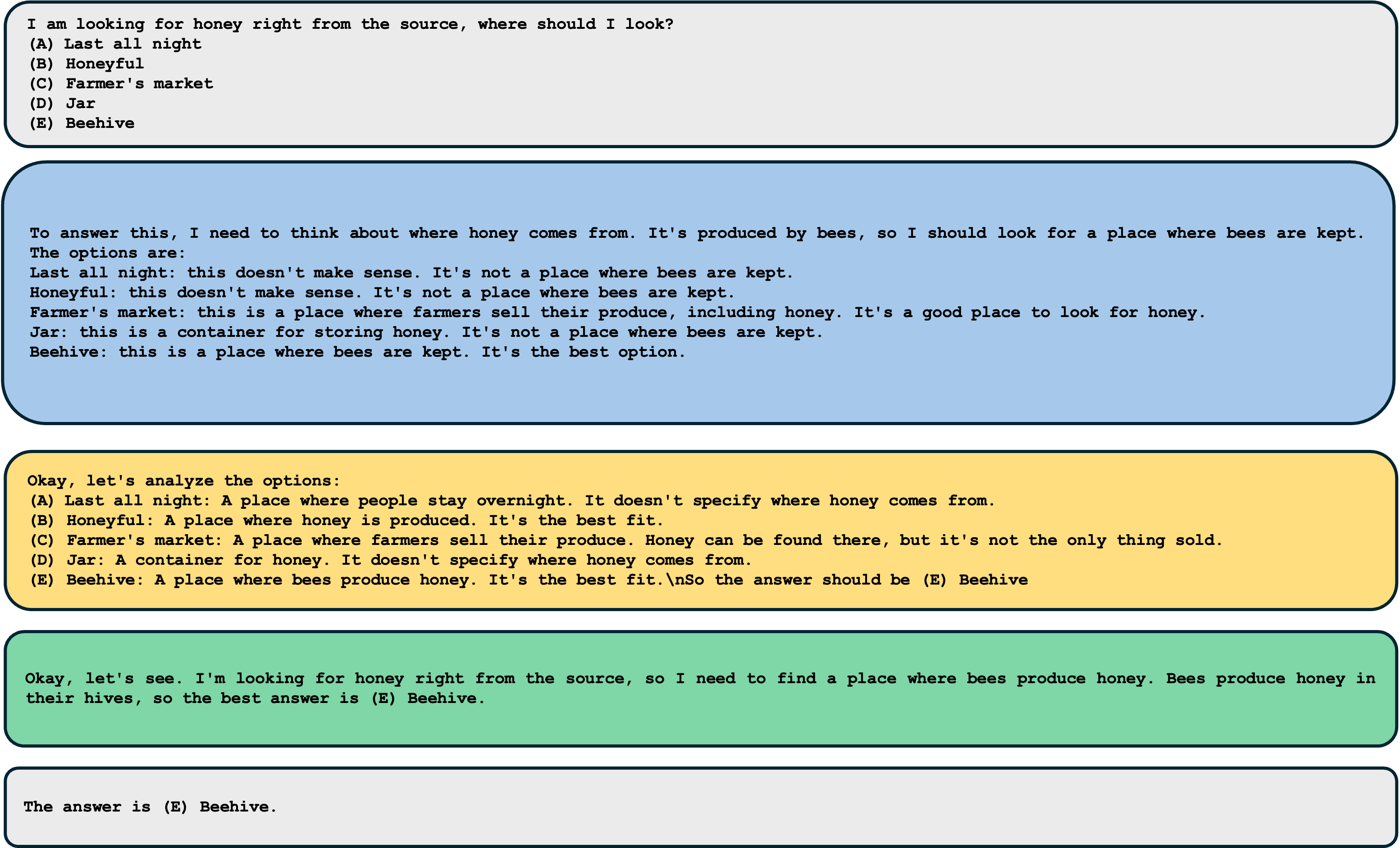}
    \caption{Qualitative examples of reasoning processes for CommonsenseQA dataset}
    \label{fig:examples_cqa}
\end{figure}

\begin{figure}
    \centering
    \includegraphics[width=\linewidth]{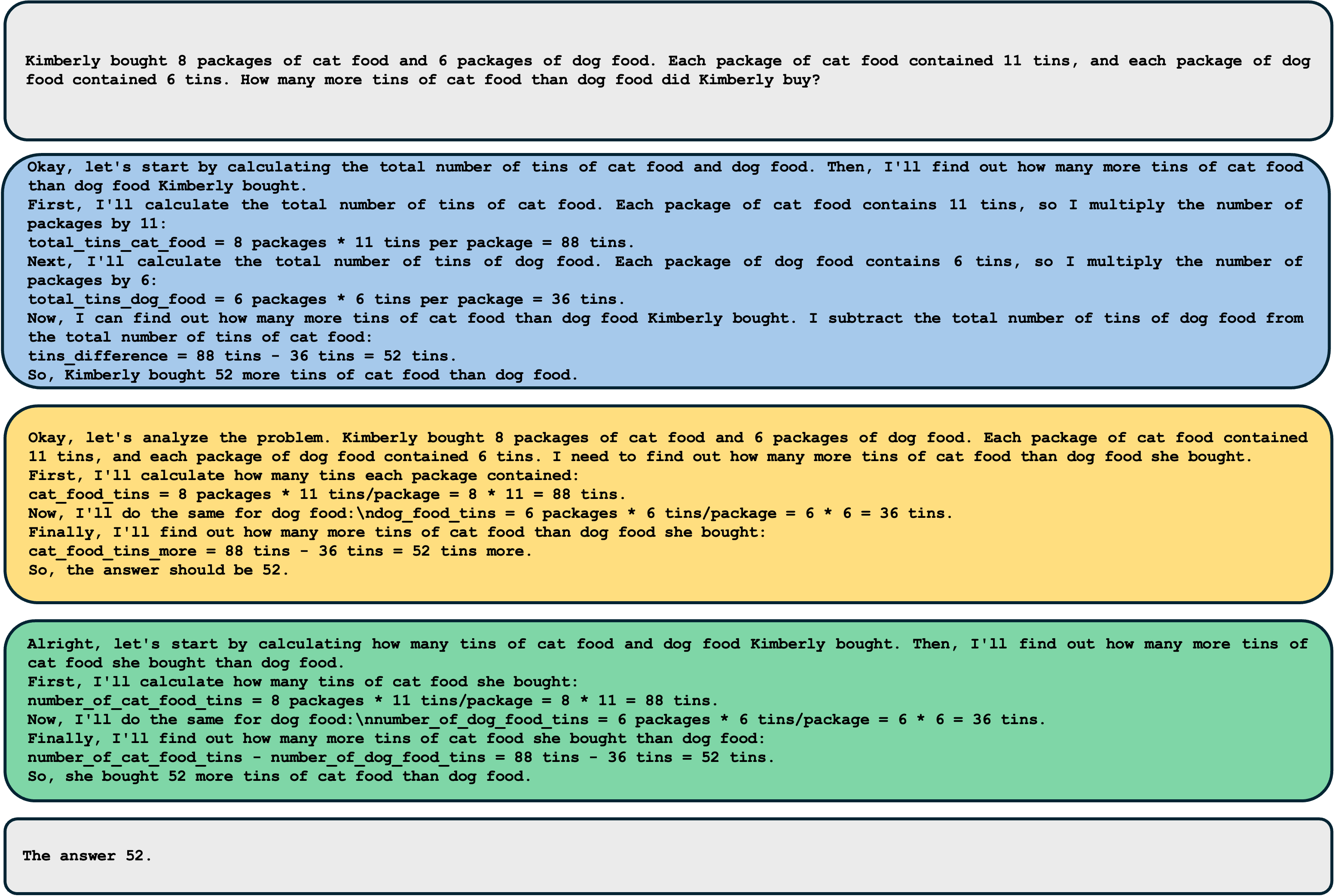}

    \caption{Qualitative examples of reasoning processes for GSM8k dataset}
    \label{fig:examples_gam8k}
\end{figure}

\begin{figure}
    \centering
    \includegraphics[width=\linewidth]{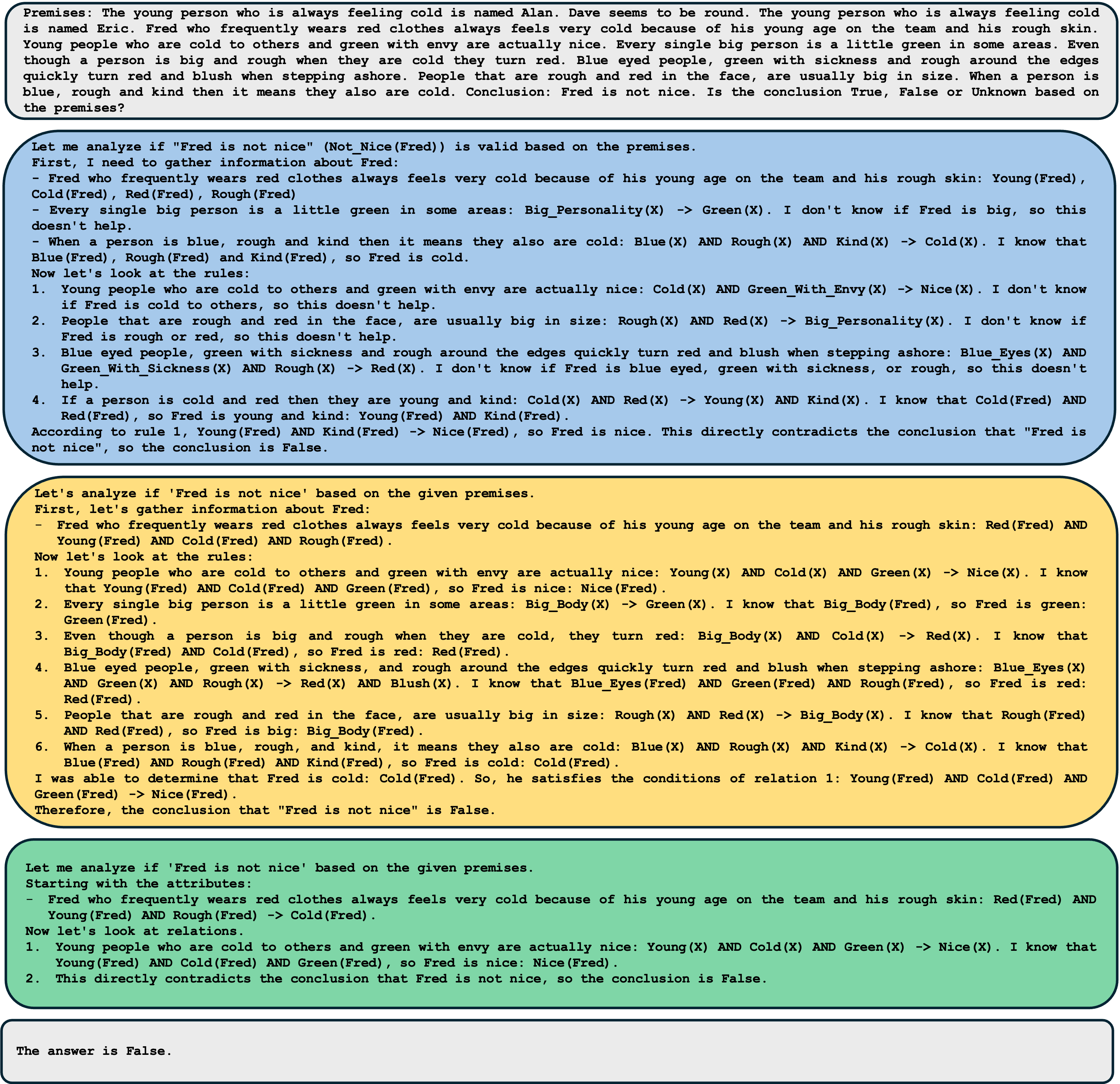}
    
    \caption{Qualitative examples of reasoning processes for Proofwriter dataset}
    \label{fig:examples_pw}
\end{figure}

\end{document}